\documentclass[10pt,twocolumn,letterpaper]{article}

\usepackage{todonotes}
\usepackage[camera]{cvpr}      % To produce the REVIEW version
\definecolor{cvprblue}{rgb}{0.21,0.49,0.74}
\usepackage[pagebackref,breaklinks,colorlinks,citecolor=cvprblue]{hyperref}

\usepackage{multirow}
\usepackage{amsthm,amsmath,amssymb} 
 
\usepackage{subcaption}

\theoremstyle{definition}

\usepackage{booktabs,xcolor,colortbl}
\usepackage{xparse}

\ExplSyntaxOn
\NewDocumentCommand{\longdash}{ O{2} }
 {
  --\prg_replicate:nn { #1 - 1 } { \negthinspace -- }
 }
 
\ExplSyntaxOff

\newcommand{\s}{\;}
\newcommand{\cwm}{\ensuremath{\mathcal{C}\mathcal{W}\mathcal{M}}}
\newcommand{\cwms}{\ensuremath{\mathcal{C}\mathcal{W}\mathcal{M}s}}
\newcommand{\simscore}{\ensuremath{{ss}}}
\newcommand{\nhl}{\ensuremath{\mathcal{L}}}
\newcommand{\nhn}{\ensuremath{\mathcal{H}}}

\newcommand{\dis}{\ensuremath{\mathcal{D}\mathcal{S}}}

\newcommand{\simmap}{\ensuremath{\mathcal{S}\mathcal{M}}}
\newcommand{\cnn}{\ensuremath{f}}
\newcommand{\latent}{\ensuremath{z}}
\newcommand{\mlp}{MLP}

\newcommand{\convWeights}{\ensuremath{%\mathcal
{W}^{conv}}}

\newcommand{\superprototype}{\ensuremath{\mathcal{S}\mathcal{P}}}

\newcommand{\inputimage}{\ensuremath{x}}

\newcommand{\prototype}{\ensuremath{\mathcal{P}}}

\RequirePackage{fix-cm}
\newcommand{\size}[2]{{\fontsize{#1}{0}\selectfont#2}}

\title{ProtoArgNet: Interpretable Image Classification with \\ Super-Prototypes and Argumentation}

\author{Hamed Ayoobi\\
Imperial College London\\
United Kingdom\\
{\tt\small h.ayoobi@imperial.ac.uk}
\and
Nico Potyka\\
Cardiff University\\
United Kingdom\\
{\tt\small potykan@cardiff.ac.uk}
\and
Francesca Toni\\
Imperial College London\\
United Kingdom\\
{\tt\small f.toni@imperial.ac.uk}
}

\begin{document}
\maketitle
\begin{abstract}
  We propose \emph{ProtoArgNet}, a novel interpretable deep neural architecture for image classification in the spirit of prototypical-part-learning as found, e.g., in ProtoPNet. While earlier approaches associate every class with multiple prototypical-parts, ProtoArgNet uses \emph{super-prototypes} that combine prototypical-parts into a unified class representation. This is done by combining local activations of prototypes in an MLP-like manner, enabling the localization of prototypes and learning (non-linear) spatial relationships among them. By leveraging a form of \emph{argumentation}, ProtoArgNet is capable of providing both supporting (i.e. `this looks like that') and attacking (i.e. `this differs from that') explanations. We demonstrate on several datasets that ProtoArgNet outperforms  state-of-the-art prototypical-part-learning approaches. Moreover, the argumentation component in ProtoArgNet is customisable to the user's cognitive requirements by a process of sparsification, which leads to more compact
 explanations compared to state-of-the-art approaches.
% ProtoArgNet addresses the problem of missing geometrical correlations, a feature that CNNs have but prototype learning approaches miss.  
%Experiments show that ProtoArgNet outperforms other methods in terms of classification accuracy and is able to encode geometrical relation in the input. Moreover, it provides an end-to-end architecture to learn an argumentation framework for the problem of image classification.

\end{abstract}    
\section{Introduction}
\label{Introduction}

% \todo[inline]{desiderata for interpretability: transparent model ??? (like all prototype-based models?)...mechanistic interpretability....
% cognitive tractability (small size) and conceptual interpretability (each component has a meaning...)}
% \todo[inline]{Change the terminology of supported and attacked regions for super-prototypes.}
\noindent Deep neural architectures are successful in various tasks \cite{lecun2015deep}, including image classification (the focus of this paper). However, 
they
% deep neural architectures for image classification (e.g. convolutional neural networks) 
tend to be mostly inscrutable black-boxes. 
In high-stakes settings, 
% instead, 
interpretability is crucial and
% , arguably,  
interpretable models %should be favoured 
are advocated, especially if they achieve comparable performance
% former exhibit acceptable performances in comparison with the latter
~\cite{Rudin19}.    %\todo{throughout intro, refer to sections rather than subsections?}  

\begin{figure}[t]
    \centering
    \includegraphics[width=1\linewidth]{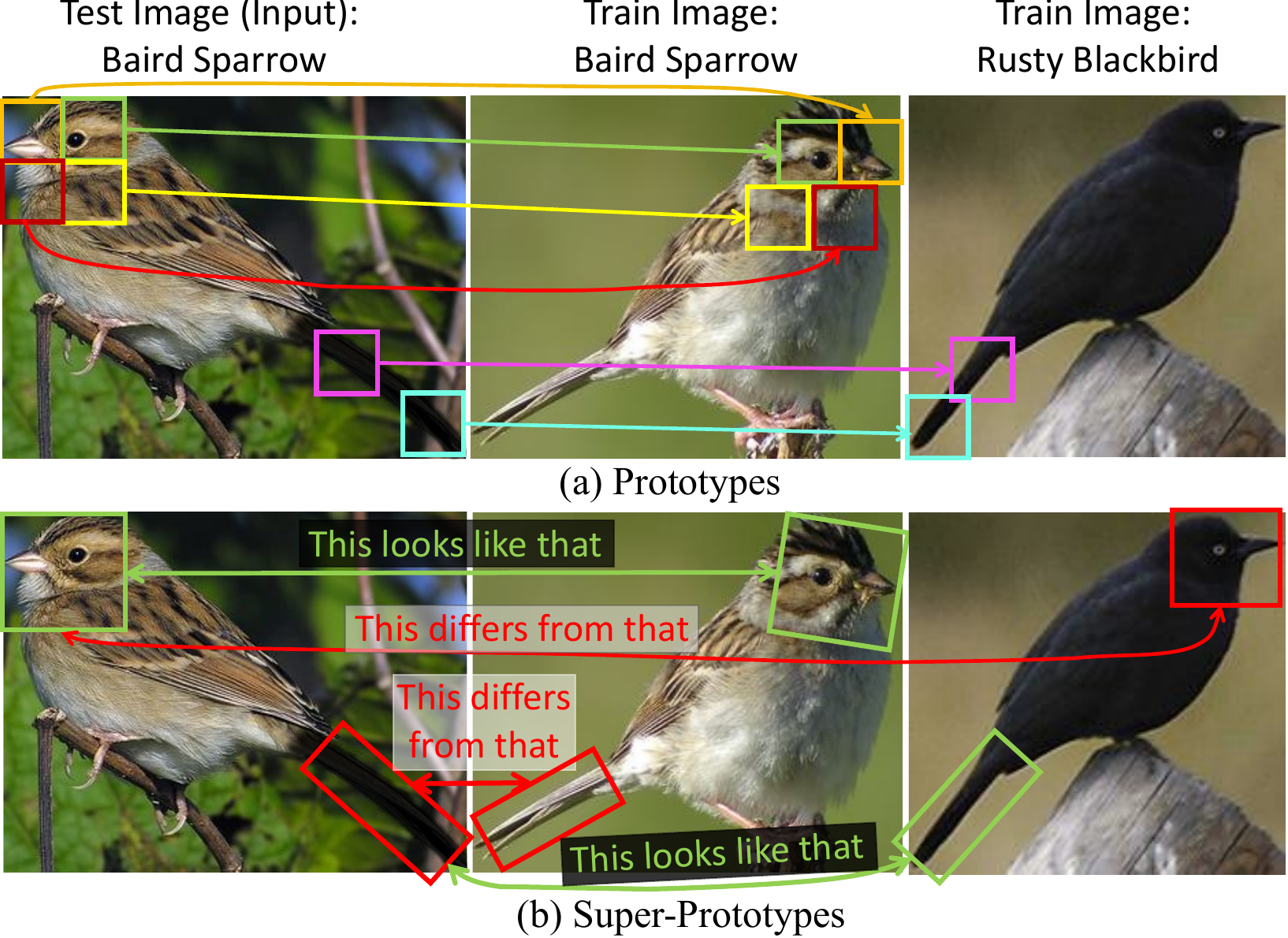}
    \caption{Conventional prototypes (a) versus the proposed super-prototypes (b) for a test image in the CUB dataset~\cite{CUB_200_2011} with the tail intentionally coloured black. Class-specific super-prototypes encode spatial correlation between prototypical-parts by combining the low-level prototypes. They provide both `this looks like that' and `this differs from that' explanations.
    }
    \label{fig:protoVSsuperProto}
\end{figure}

Prototypical-part-learning for image classification  %\todo{add a reference? }
amounts to learning proto\-typical-parts of classes in images by introducing a \emph{prototype layer}
between a \emph{convolutional backbone} and a \emph{classifier} \cite{chen2019looks}. 
% Intuitively, 
Prototypical-parts
% correspond to
are latent representations of patches %\todo{Is this well known as a term? why not just patches? Yes, I think it is a standard term.}
in images%. Intuitively, they represent meaningful patches of images
, like the beak or tail of a bird (see Figure~\ref{fig:protoVSsuperProto} (a)).
The prototype layer determines the similarity between prototypical-parts and patches in the latent space that the convolutional
backbone maps to. 
While some prototypical-parts may %correspond to meaningful parts in the input image, they may also 
% correspond to %background 
% patches that are 
be meaningless for humans, 
% (rather than exclusively meaningful parts in images as in Figure~\ref{fig:protoVSsuperProto} (a)),
the same can be said about some of the
latent features learnt by black-box models
\cite{jo2017measuring}.
The transparency of protypical-part approaches
allows detecting if a decision has been
made based on meaningful patterns or
statistical artefacts.
% they allow making transparent
% classifications, based on clearly defined prototypes,
% % insofar as
% if the classifier is interpretable (e.g., logistic regression as in ProtoPNet \cite{chen2019looks}). 

\begin{figure*}
    \centering
    \includegraphics[width=0.98\textwidth]{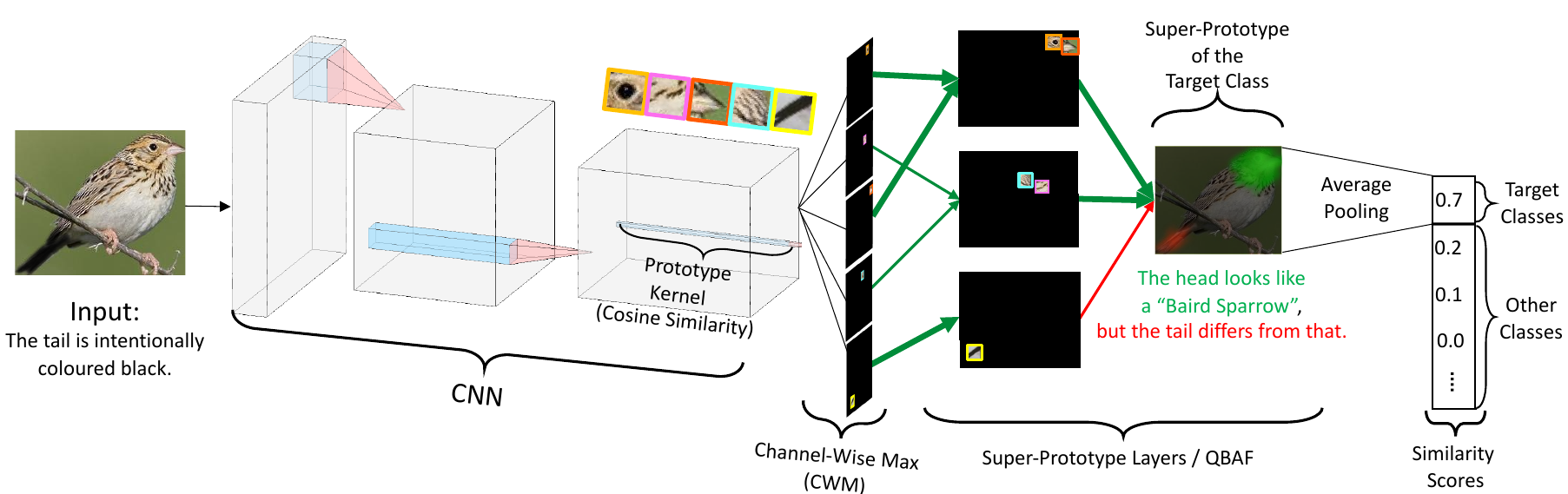}
    \caption{Architecture of ProtoArgNet (details in  Section~\ref{sec:ProtoArgNet}), illustrated with a sample from the CUB dataset with the tail intentionally coloured black.}
    \label{fig:architecture}
\end{figure*}

\begin{figure}
    \centering
    \vspace{-3mm}
    \includegraphics[width=1\linewidth]{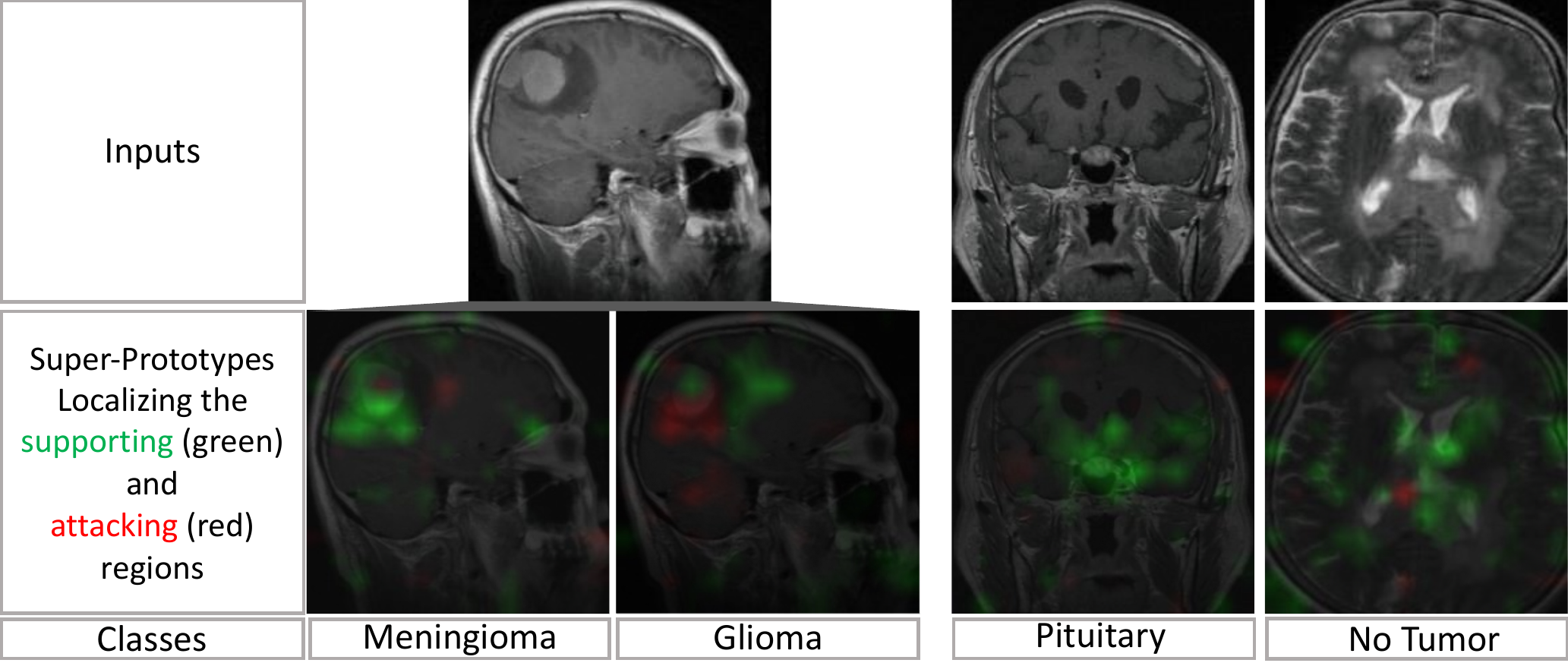}
    \caption{%Classification of 
    Sample inputs from the Brain Tumor MRI dataset~\cite{brainMRIdataset} (top row), and 
    corresponding super-prototypes by ProtoArgNet (bottom row), localizing the regions supporting (green overlay) and attacking (red overlay) the corresponding classes (details in Section~\ref{Experiments}).}
    \label{fig:brain}
\end{figure}

We propose \emph{ProtoArgNet} (Section~\ref{sec:ProtoArgNet}, overviewed in Figure~\ref{fig:architecture}), a novel interpretable deep neural  architecture for image classification 
in the spirit of prototypical-part-learning.
Similar to ProtoPShare \cite{RymarczykST021} and ProtoTrees \cite{NautaBS21}, 
% To obtain super-prototypes, in 
ProtoArgNet shares 
prototypes among classes. 
% layer is not divided amongst separate classes, and all classes share the same prototypes. 
% To group prototypes,  ProtoArgNet uses cosine similarity (rather than a translation of L2 distance into similarity scores as in ProtoPNet), easily computed by a normalized convolutional layer. 
% To obtain super-prototypes from prototypes, 
% ProtoArgNet uses a channel-wise max layer followed by the super-prototype kernel to learn a single super-prototype for each class. 
However, 
while %earlier 
%these and other
existing
prototypical-part-learning approaches associate every class with multiple prototypical 
parts,
ProtoArgNet 
summarizes them in a single \emph{super-prototype} per class.
Intuitively, the super-prototype combines
local activations of prototypes to encode spatial relationships %between 
amongst them
% uses  
% \emph{super-prototypes}, a compact encoding of all the prototypes for each class combining spatially correlated prototypical-parts into single prototypical class representations
~(see Figure \ref{fig:protoVSsuperProto} (b) for an 
illustration).
% By  using a single super-prototype per class instead of thousands of prototypes as in
% state-of-the-art prototypical-part-learning approaches (notably 
% ProtoPNet \cite{chen2019looks}, ProtoTrees \cite{NautaBS21}, ProtoPShare \cite{RymarczykST021}, and PIP-Net \cite{Nauta23}),
% ProtoArgNet 
% supports interpretability without the need for pruning as a post-processing step to reduce the number of prototypes.
%
%
%
As we will demonstrate in the experiments with the SHAPES dataset~\cite{SHAPES},
% Specifically, super-prototypes can encode  spatial relationships amongst prototypical object parts which 
these relationships are essential for some classification tasks 
but state-of-the-art prototypical-part-learning approaches 
are unable to capture them.
% (whereas CNNs do, by capturing relations between patterns recognized in earlier layers). 
This localization of prototypical-parts can
be particularly useful in medical diagnosis \cite{xprotonet} where the model can predict the location of disease indicators without requiring masks for the training data. Figure \ref{fig:brain} shows an application of ProtoArgNet to
an MRI scan for brain tumor diagnosis \cite{brainMRIdataset}.

\setlength{\intextsep}{5pt}%
%Unlike CNNs, prototypical-part-learning approaches are unable to reason based on the location of prototypical-parts. The reason is that the prototype layer in these approaches generally provides information about the presence of prototypical-parts in the input image and they can not provide information about the spatial relationship between the prototypical-parts. 

% For example, in Figure \ref{fig:SHAPES}, a positive
% example (Class $1$) has a triangle in the left column and a circle in the right column on the same row. Merely recognizing prototypical-parts for triangles and circles in the input image (as in other prototypical-part-learning approaches) is insufficient for determining the class label in this example.

% As opposed to other prototypical-part-learning approaches,
The super-prototype layers in 
ProtoArgNet can
capture non-linear relationships, similar to a Multi-Layer Perceptron (MLP). 
However, instead of operating on individual
neurons, ProtoArgNet operates on activation maps (Section~\ref{sec:ProtoArgNet}).
Since MLPs can, in particular, learn logical functions like disjunction and XOR, ProtoArgNet can also learn classes
that cannot be captured by atomic 
spatial patterns (Section~\ref{Experiments}). %HA: too complex for the starting point and it needs supporting experiments in the main body of the paper.  
%For example, while a single activation map could only capture that the tail of a bird must be to the right of its head, a second map could capture the pattern that the tail is on the left of its head (which corresponds to a 180 degree rotation). However, to capture increasingly complex pattern, the network has to become more complex.
To address the lack of interpretability of large  MLPs, ProtoArgNet applies 
the SpArX methodology of \cite{AyoobiPT23} to translate the MLP to a %sparse
\emph{quantitative bipolar argumentative framework} (QBAF)~\cite{Potyka_21},  
a well-known form of \emph{argumentation}~\cite{AImagazine17}. The `Arg' in ProtoArgNet refers to the use of QBAFs. ProtoArgNet is customisable to user cognitive requirements by sparsifying the MLP/QBAF component. 
The sparse QBAF explains the mechanics of 
the underlying MLP in terms of the roles played by the prototypes towards super-prototypes through the hidden (clusters of) activation maps. In the QBAFs, 
 the `arguments' (amounting, in ProtoArgNet, to channel-wise maxes, clusters of hidden activation maps in the MLP, and super-prototypes) can `attack' or `support' other `arguments' (as indicated with red and green arrows in Figure~\ref{fig:architecture}), with a dialectical strength in line with activations in the MLP.

In summary, we make the following main contributions:
\begin{itemize}
    \item We propose \emph{super-prototypes}, which are class-specific combinations
    of prototypical-parts
    % . Unlike existing prototypical-part-learning approaches, it allows the localization of prototypical-parts and 
    that allow capturing
    spatial relationships between them.
    % The super-prototypes are class-specific and cognitively more tractable.
    %ProtoArgNet effectively tackles this challenge by encoding the spatial relations between distinct prototypical-parts using the super-prototype kernels.
    \item We present \emph{ProtoArgNet}, a novel prototypical-part-learning approach integrating super-prototypes 
    and QBAFs for improved performance and interpretability.
    \item 
We show experimentally that ProtoArgNet 
outperforms 
the state-of-the-art prototypical-part-learning models ProtoPNet \cite{chen2019looks}, ProtoTree \cite{NautaBS21}, ProtoPShare \cite{RymarczykST021}, ProtoPool \cite{protoPool} and PIP-Net \cite{Nauta23} in terms of classification accuracy, explanation complexity, and the ability to encode and detect (non-linear) spatial relationships in images.
% , supported by an ablation study

%\todo{The footnote may be a bit confusing. We will make the code available in the supplementary material, right? Then perhaps just write, we will add a URL to the code after acceptance} 
% and the study of the cognitive complexity and cognitive tractability of local explanations derived from the sparsification of QBAFs obtained with ProtoArgNet.
\end{itemize}

\section{Related Work}
\label{Related Work}

The problem of explaining the outputs of image classifiers is well-studied in the literature.
Post-hoc explanation approaches like
feature attribution methods  \cite{LIME_2016should,SHAP,deeplift}, attention maps \cite{integratedcam}
or counterfactual explanations 
\cite{goyal2019counterfactual}) 
%neuron level inspection (e.g. see \cite{olah2017feature,olah2020zoom}) or
% and concept-based methods (e.g. as in \cite{tace,dissect%,concepttrail
% }).
aim at explaining black-box models.
% These approaches fall under the category of \emph{post-hoc explanations} for the outputs of
%  models, specifically to be used when the models are  black-boxes that need to be understood.
We focus instead on developing an \emph{interpretable}  model based on \emph{prototypical-part-learning}~\cite{snell2017prototypical}
 and \emph{argumentation}~\cite{AyoobiPT23}. 
Prototypical-parts have been
introduced in
\emph{ProtoPNet} \cite{chen2019looks}.
% introduced prototypical-parts %that are supposed to capture SEEMS NEGATIVE - THEY ARE SUPPOSED TO BUT ARE THEY?
% for capturing
% parts of a class (like the beak or tail of a bird) rather than the whole object (the bird).
% and showed that it 
% performs competitively on %Birds
% the more demanding CUB-200-2011 (CUB)~\cite{CUB_200_2011} %\todo{WHY NOT SAY CUB? DO THEY JUST FOCUS ON BIRDS? CUB is only about birds. Probably Nico wanted to mention the use case and not the dataset names?! }
% and Stanford Cars (Cars) \cite{CARS} datasets.
ProtoPNet learns prototypical-parts as subpatches of the output of a 
convolutional backbone. 
A prototype layer
associates each class with $m$ prototypes and 
determines the maximum similarity 
% (inverse Euclidean distance) 
between patches in the input image and prototypes. % and returns, for each class, the maximum similarity to a prototype of this class
The classification is then made by logistic regression 
based on the individual similarity values.
% \iffalse
% As regards the logistic regression classifier, it is trained in two stages: first the regression weights for prototypes are fixed (to  1 if the prototype is associated to classes and -0.5 else) while the convolutional backbone and prototype layer are trained;  
% %Classification is again done by logistic regression on the outputs of the prototype layer. In the first stage of the training process, the regression weights for each class are set to $1$ for the associated  prototype and to $-0.5$ for all other prototypes. The convolutional and prototype layer are then trained while the regression weights remain fixed. 
% then 
% %In the second stage, 
% each prototype is replaced with the closest convolutional output from the same class in the training data to make sure that prototypes correspond to actual parts of images; finally, with the weights of the convolutional backbone and prototype layer fixed, the regression weights are optimized. \citeauthor{chen2019looks}\cite{chen2019looks} demonstrate
% on the Birds~\cite{CUB_200_2011} %\todo{WHY NOT SAY CUB? DO THEY JUST FOCUS ON BIRDS? CUB is only about birds. Probably Nico wanted to mention the use case and not the dataset names?! }
% and Cars \cite{CARS} datasets that ProtoPNet can generate intuitive explanations, while the classification performance is not too far
% from state-of-the-art black-box models.
% \fi
ProtoPNet has been extended in different directions.
\emph{ProtoPShare} \cite{RymarczykST021} improves ProtoPNet 
by sharing prototypes among classes. 
% To this end, prototypes are pruned based on
% a {data-dependent similarity} between prototypes that
% measures the compliance on similarity scores for all patches
% in the training data. Similar prototypes are merged and can be 
% shared between classes. Experiments %\todo{on which datasets? on the same datasets that are used for ProtoPNet which are CUB and Cars} 
% (again with CUB and Cars) show that ProtoPShare can
It can achieve the same performance as ProtoPNet while reducing the number
of prototypes by 50-75\%. 
\emph{ProtoTree}~ \cite{NautaBS21} builds on ProtoPNet by using a soft decision 
tree, rather than logistic regression, on top of the convolutional backbone.
% Every node in the tree learns a prototype similar to ProtoPnet's
% prototype layer.
%Given an input, nodes assign probabilities to their two children  based on the input's similarity to the prototype. Leaves learn a class distribution and the final class distribution generated by ProtoTree is a convex combination of the leaves' distributions weighted by  the probability of reaching the leaf. 
% While the trees grow exponentially %with their 
% in height,  %\citeauthor{NautaBS21} demonstrate that the trees
% they 
% can be pruned significantly and ProtoTree can achieve 
Prototypical-parts occur now in decision nodes,
which results in increased performance 
% than ProtoPNet (again on CUB and Cars) 
while reducing the number of prototypes 
 by 90\%. %Experiments %\todo{on which datasets? On exactly the same datasets as ProtoPNet} 
 %also indicate that the %soft
%  trees
% can be made deterministic without harming classification
% performance significantly.\todo{do we really care about this last bit? I don't think so. I will remove it, then.}
% \cite{hoffmann2021looks} noted
% that similarity in
% the latent space does not necessarily correspond to similarity
% in the image space.
% This makes prototypical-part-learning approaches
% prone to adversarial manipulations 
% resulting in explanations that
% are absurd for humans.
\emph{PIP-Net} \cite{Nauta23} aims to address
the problem that similarity in
the latent space does not necessarily correspond to similarity
in the image space \cite{hoffmann2021looks}
by applying data augmentation and a new 
alignment term.
% this issue by first creating pairs of
% images that share prototypes from single images by applying 
% data augmentations
% and then adding an alignment term to the training objective
% that penalizes prototypes that are dissimilar for image pairs. 
% \todo{actually, this can solve only half of the problem.
% The loss term encourages that images with the same prototypes
% will be similar for prototypes in the image space,
% but the converse problem that similarity in the image space
% does not entail similarity in the image space remains unaddressed}
% deformable prototypes do not necessarily seem relevant here. 
% the results in the paper are also a bit weak and the idea
% seems rather random (instead of dividing one big prototype
% into multiple smaller parts, one may as well use several
% small prototypes in the first place - this is indeed what seems
% to happen in most papers as they use 1x1 prototypes that 
% cannot even be "deformed"
% \cite{DonnellyBC22}
% \cite{BontempelliTTGP23}
% While these approaches need to integrate a step to prune a large number of trained prototypes, 
\emph{ProtoPool} adds fully differentiable
assigment of prototypes to classes during 
training using %the soft assignment of prototypes with 
the Gumbel-Softmax trick~\cite{gumbelSoftmax}, while reducing the required number of prototypes by sharing prototypes across all classes, similarly to ProtoPShare. Further,  ProtoPool uses a focal similarity function
to distinguish prototypes from less salient, background features.
%, concentrating on more salient visual features.
 ProtoArgNet differs from these state-of-the-art approaches in that it uses super-prototypes and MLPs/QBAFs, based on a novel architecture. 
 % We will use ProtoPNet, ProtoPShare, ProtoPool, ProtoTree and PIP-Net as baselines to evaluate ProtoArgNet's performance. 

% \subsection{Argumentation}
%\todo[inline]{1. How much do we want to stress the argumentation side? (Write generic about Argumentation Based Explainability). \\ 2. We can mention that there is only limited work about argumentation and images } 

ProtoArgNet uses a form of argumentation \cite{AImagazine17},  to explain super-prototypes.
% (lost with the use of MLPs rather than logistic regression or other interepretable methods in the state-of-the-art on prototypical-part-learning). 
Speficially, ProtoArgNet extends the \emph{SpArX} approach~\cite{AyoobiPT23}, originally defined for MLPs with tabular data only, to the setting of prototypical-part-learning with images.
Several  argumentation-based forms of explainability have been proposed in recent years \cite{argXAIsurvey}. 
% The use of argumentation within ProtoArgNet paves the way to customizing ProtoArgNet to generate a variety of explanation formats. While doing so it outside the scope of this paper it is an important avenue for future work.
Other works combine argumentation and image classification, e.g. \cite{DAX,LRPArgExpl21} for explaining the outputs of CNNs   and \cite{Ay:AABL} to obtain an interpretable image classifier%; potentially Avinash' work on visual debates? (but it is just an arxiv for now) -- explainining CNNs by a surrogate model based on quantization and debates
% . We focus on combining argumentation and the novel concept of super-prototypes for interpretability purposes%, in line with SpArX, but for images
. 
To the best of our knowledge, ProtoArgNet is the first approach to use argumentation for prootytpical-part-learning.

\section{Preliminaries}
\label{Preliminaries}
% \todo[inline]{Hamed: Example of What Sparx is doing with an illustration using Figure 1 in the SparX paper. We should write that we have borrowed the figure with their permission. }
We build up on SpArX \cite{AyoobiPT23}, 
a post-hoc explanation method
that aims at generating structurally faithful explanations for
MLPs. 
SpArX exploits that MLPs can be
understood as a special case of Quantitative Bipolar Argumentation Frameworks (QBAFs)
\cite{Potyka_21}. QBAFs are graphical reasoning models, where %abstract 
nodes represent
\emph{abstract arguments} 
and edges represent \emph{attack} or \emph{support} relations between the arguments.
% , each with a
% (negative or positive)
% intensity value.
%
Every argument in a QBAF
is associated with an \emph{initial strength} and reasoning algorithms determine a
\emph{final strength} (representing an acceptability degree) for every argument, based on its
initial strength and the final strength of its attackers and supporters.
% Formally, the final strength values are a fixed-point of an update
% function.

Arguments in QBAFs are abstract entities. What makes them arguments is that they are in dialectical relationships with each other. 
% To capture MLPs as in \cite{Potyka_21}, these abstract arguments represent input features,  hidden neurons and output classifications, and the graphical structure of the QBAF mirrors the one of the MLP. 
%
Roughly speaking, in order to transform an MLP into a QBAF, 
neurons can be associated with %abstract
arguments,
their biases can be transformed into initial strength values and
their connection weights into intensity values of attack and support relations. 
The translation guarantees that the activations of neurons in the original MLP correspond to 
the final strength values of arguments in the QBAF under particular semantics \cite{Potyka_21}. 
While this correspondence allows representing MLPs faithfully by QBAFs,
it does not add much interpretability
% and explainability, 
because the QBAF 
has the same size as the original MLP. Thus, SpArX sparsifies the network by 
clustering nodes with similar activations and representing each
cluster by a single argument~\cite{AyoobiPT23}.
% Naturally, a larger compression ratio (sparser graph) can result in larger unfaithfulness 
% (average difference between the strength of the cluster argument and the activations of the neurons in the cluster). 
% Experiments with tabular data show that
% SpArX can give explanations that are both sparse and faithful. 
% The sparsification process is 
% designed in SpArX maintains classification accuracy for local explanations, as the obtained QBAF remains faithful to the original model~\cite{AyoobiPT23}.

In this work, we extend SpArX to
make  ProtoArgNet interpretable and explainable. 
An illustration is given in Figure~\ref{fig:architecture}:  activation maps
% \todo{do people know what is meant by channel? Couldn't we say something like "activation maps" which seems more telling to me? What is confusing is that "channel" is often used for scalar values in other contexts. I suppose the motivation here comes from "color channels" in computer vision, but not sure if this is clear for the reader outside of computer vision. HA: You are right. It might be more common for the computer vision community. I've changed it to activation maps in the paper. }
in the super-prototype layers of ProtoArgNet are treated as arguments, alongside the Channel-Wise Maxes (CWMs) that localize the prototypes, which serve as the input features for the super-prototypes layers in our architecture, as presented next. 
%(Section~\ref{sec:ProtoArgNet}).
% Similarly to the original SpArX, we experiment with sparsification by various compression ratios ( Section~\ref{Complexity}), showing that ProtoArgNet can provide explanations that are both sparse and faithful for image classification.

% \iffalse In this work, we extend SpArX to
% THE BELOW IS A REPETITION
% convolutional neural networks (CNNs)
% by building up on the prototypical-part idea from ProtoPNet \cite{chen2019looks}. While classical CNNs can be divided into a convolutional block followed by a classification block, ProtoPNet and its successors add a prototype block in between. The classification block uses the outputs of the prototype block
% rather than the output of the convolutional block. In case of ProtoPNet \cite{chen2019looks} and ProtoPShare \cite{RymarczykST021}, the prototypical block is an independent layer that is followed by logistic regression for classification. In case of ProtoTree \cite{NautaBS21}, the classifier is a decision
% tree and the prototypical block is integrated into the classifier by associating
% decision nodes with prototypical-parts. Our architecture is closer to ProtoPNet and ProtoPShare in that the prototype block is an independent block between
% the convolutional and classification block, but it is more sophisticated
% as we will explain later.
% We will also replace the logistic regression classifier with a QAF that is 
% generated from an MLP following the SpArX methodology \cite{AyoobiPT23} described above. 
% \fi

\section{ProtoArgNet}
\label{sec:ProtoArgNet}
% \todo[inline]{Start with high level overview to mention the parts of the architecture. Then, explain intiuitively what each part is doing. Finally, give formal definition (techinical) for each part.}
Figure \ref{fig:architecture} shows the architecture of ProtoArgNet. ProtoArgNet consists of a %CNN 
convolutional backbone \cnn\ with weights \convWeights%\todo{where are these weights used? much later...}
, a prototype layer \prototype, a Channel-Wise Max (\cwm) layer,  and a Super-Prototype layer
%kernel 
\superprototype\ mapped onto a QBAF for interoperability and explainability purposes.
 % followed by an MLP \mlp\s with weights $W^{\mlp}$, %which will be translated to 
 % mapped onto a %Quantitative Bipolar Argumentation Framework (QBAF) 
 % QBAF for interpretability and explainability purposes. 
We discuss each component in turn, assuming that inputs are images and the classification task amounts to predicting a class in the set $K$ ($|K|\geq 2$).

% \todo[inline]{start every subsection with a high-level description of what this part of the architecure is supposed to do intuitively}

\subsection{Prototypes}\label{sec:prototypes}
% \todo[inline]{what is this part supposed to do and why is it 
% defined in this particular way? Hamed: what do we formally mean by prototypes? Explain what prototypes are. We can refer to the background as well. 
% Include definitions for each of the equations here. 
% }

Let $\latent=f(\inputimage)$ be the convolutional output %of the CNN 
for an input image $\inputimage$, where  the output tensor $\latent$ has shape $H \times W \times D$  with height $H$, width $W$ and $D$ channels. This output tensor serves as input to the prototype layer, \prototype.
%The prototype layer 
which represents prototypical-parts. 
\prototype~consists of $N$ prototypes $P = \{p_i\}_{i=1}^{N}$ with shapes $H_1\times W_1\times D$. 
% \todo[inline]{explain the choice of H and W in the experiments}
As usual, we use $H_1 = W_1 = 1$. 
% \todo[inline]{what do you mean by $\latent_j \in \latent$? 
% I would expect that $\latent$ is a tensor, not a set.  }
For each prototype $p_i \in %\prototype
P$ and every $1\times 1\times D$ sub-tensor $\latent_{h,w,.}$ of  $\latent$, the prototype layer \prototype\ computes the cosine similarity 
$\simmap^i_{h,w,.} = \frac{p_i \cdot \latent_{h,w,.}}{\|p_i\|\|\latent_{h,w,.}\|}$
% \begin{equation}
%     \csn(p_i, \latent_j) = 
% \end{equation} 
%The prototype layer 
and summarizes the similarity values in a matrix 
$\simmap^i$ of
dimension $H \times W$.
% That is, $\simmap^i_{h,w,.} = \frac{p_i \cdot \latent_{h,w,.}}{\|p_i\|\|\latent_{h,w,.}\|}$.
% \begin{equation}
%     \simmap^i = \underset{\latent_j \in \latent}{\csn}(p_i, \latent_j) 
% \end{equation}
% with shape $H \times W$ for each prototype $p_i \in P$. 
Intuitively, a similarity map $\simmap^i$ indicates
how similar the prototypical-part $p_i$ is to patches of
the input image $\inputimage$ in the latent space. 

Compared to the commonly used approach of computing L2 distance and converting it to similarity (as in ProtoPNet, ProtoPShare, and ProtoTrees), cosine similarity is scale-invariant and thus more easily interpretable.
% , and more computationally efficient.  
% \begin{definition} 
% \end{definition}
% Each $z_j$ in the convolutional output corresponds to an image patch in the input image $x$ considering the receptive field of the convolutional backbone. 
We implemented $\simmap$ using the 2D convolution operator $\ast$. It generates $\simmap^i$ by convoluting the normalized convolutional output $\hat{\latent} = \frac{\latent}{\|\latent\|} = \big[\frac{\latent_j}{\|\latent_j\|}\big]_{\latent_j \in \latent}$ with a normalized prototype kernel $\hat{p_i} = \frac{p_i}{\|p_i\|}$,  $\simmap^i = \hat{\latent} \ast \hat{p_i}$. 
% The similarity map $\simmap^i$ determines to what extent the prototype $p_i$ looks like the convolutional outputs $\latent_j \in \latent$ in each image patch. 
Since cosine similarity is used for the prototype layer, the values in similarity maps can be both positive and negative in the range $[-1, 1]$. The output dimensions of the prototype layer are $H \times W \times N$%, where $N$ is the number of prototypes. SAID ALREADY
%\todo{is $N$ learnt or a hyper-parameter? A hyper-parameter that is mentioned in Section 6.1}
.

\subsection{Channel-Wise Max} \label{sec:cwm}
% \todo[inline]{motivate why before explaining how}
The Channel-Wise Max layer aims to localize and extract the maximum value from each similarity map, while ensuring that only one prototype is activated at each location across all similarity maps.  
 $\cwm$ takes the similarity maps 
as input.
% and extracts the maximum value from each input channel by passing the maximum value and setting all other values to zero while preserving the input dimensions.
For each similarity map, it determines the maximal value and sets all non-maximal
values to $0$.
Formally, for every similarity map $\simmap^i$,
the channel-wise max filter creates a new map
$\cwm^i$  of the same dimension. To do so,
it
determines the
maximal value among the entries
$\simmap^i_{h,w}$,  retains the highest value $s^i_{max} = \max_{1 \leq h \leq H} \max_{1 \leq w \leq W} \simmap^i_{h,w}$ within the map and assigns a value of zero to the remaining elements, that is,
\begin{equation}\cwm^i_{h,w} = \begin{cases}
  \simmap^i_{h,w} & \text{if $s^i_{max} = \simmap^i_{h,w}$}; \\
  0 & \text{otherwise.}
\end{cases}\end{equation} 
% \begin{definition}

% \end{definition}

\noindent 
% The output dimensions of  \cwm~are still $H \times W \times N$.
It may happen that two distinct maps,
$\cwm^i$ and $\cwm^j$, have a maximal
activation at point $(h,w)$, which would
make it more challenging to interpret the subsequent layers.
To avoid this, we consider only the maximally
activated prototype at each position $(h,w)$.
To make this choice differentiable during
training, we apply the Softmax activation function \cite{softmax} to each position 
$(h,w)$ ranging over 
$\cwm^1,\dots, \cwm^N$.
% To restrict the number of activated prototypes at each of the $H \times W$ points $(h,w)$ of
% $\cwm^i$, we apply the . 
% \todo[inline]{Explain about the temperature. Separate the train and inference sentences. First about training and then about inference in the next paragraph.}
\begin{equation}\label{eq:softmax}
\cwm^i_{h,w} =  \frac{{\rm e}^{(\cwm^i_{h,w}/{T)}}} {\sum_{j=1}^{N}{\rm e}^{(\cwm^j_{h,w}/{T})}}\end{equation}
During training,  we gradually decrease the temperature parameter $T$ from $1$ to $0$. 

After training, we replace the softmax
function with the maximum
to ensure the activation of at most one prototype per location. 

\subsection{Super-Prototypes and Similarity Scores}\label{sec:super-prototypes}

The super-prototypes 
module takes the \cwms\s as input and provides a single similarity score per class. To do so, it generalizes the
mechanics of MLPs, but whereas MLPs operate on scalars, the super-prototype module operates on matrices. The input matrices are the maps $\cwm^1,\dots, \cwm^N$ and in the first layer of the super-prototype module they are combined
affinely to form new matrices of the same dimension. After applying an activation function, the matrices in this layer can then
again be combined to form matrices in the next
layer analogously to MLPs.
To describe this formally, let $A_i^l$ range over the
matrices in layer $l$. We let $A_i^0 = \cwm^i$,
and for $l>0$,
\begin{equation}\label{eq:mlp-super-prototypes}
    A_{i}^{l} = \sigma\left({(\sum_{j=1}^{N_{l-1}}{w_{ji}^{l} \cdot \; A_{j}^{l-1}}) + b_i^l}\right)   
\end{equation}
\noindent where
$N_0 = N$ is the number of prototypes and
$N_{l}$, for $l>0$, is the size of layer $l$, 
% and
% $*=\begin{cases}
%   N & \text{if \;$l = 1$} \\
%   \nhn & \text{otherwise}
% \end{cases}$, 
$b_i^l$ is a bias matrix, and $\sigma$ the activation function (GELU \cite{gelu} performed best in our experiments). 
% instead
% of combining activations at individual neurons,
% it combines the activations of map
% To do so, it forms affine combinations of the maps $\cwm^i$ and applies 
% a non-linear activation 
% This is done similarly to \mlp\s with the difference that the incoming \cwms\s are combined and aggregated. 

Like an MLP, the super-prototype layers can have various configurations regarding the number of hidden layers \nhl, the number of hidden activation maps at each layer \nhn, and the activation function $\sigma$ used at each hidden layer, hence we refer to it as Super-prototypes MLP (SMLP). 
% Unlike \mlp, the inputs are not single neurons and are the $N$ number of \cwms\s with the dimension of $H \times W$. 

% Assuming that $C_i^0=\cwm_i$, the output of a hidden channel $C_{i}^{l}$ for a hidden layer $l \in \{1, \ldots, \nhl\}$ is defined as follows: 
% % \todo[inline]{forgot to write about bias. Replace h with C.}

% \begin{equation}\label{eq:mlp-super-prototypes}
%     C_{i}^{l} = \sigma\left({(\sum_{j=1}^{*}{w_{ij}^{l} \cdot \; C_{j}^{l-1}}) + b_i^l}\right)   
% \end{equation}

The output layer provides a single super-prototype per class  $k \in K$. Each super-prototype $\superprototype_k$ is defined as follows:

    % \todo[inline]{the index $i$ of the weight is undefined. Is it supposed to be k? H: True}
\begin{equation}\label{eq:superprotoypes}
    \superprototype_k = {\sum_{j=1}^{|K|}{w_{jk}^{\superprototype} \cdot \; A_{j}^{\nhl}}}
\end{equation}

In the final step, a single similarity score $\simscore_k$ is computed for each super-prototype
by summing up the values in $\superprototype^k$:
\begin{equation}\label{eq:simscore}
    \simscore^k = \sum_{1 \leq h \leq H, 1 \leq w \leq W} \superprototype^k_{h,w}.
\end{equation}
Note that Equation \ref{eq:mlp-super-prototypes} can be efficiently implemented by employing convolutions with kernel shapes $1\times1\times N_{l-1}$, followed by an activation function. Similarly, Equation \ref{eq:superprotoypes} can be implemented using convolutions of shape $1\times1\times\nhn$.

\subsection{Super-Prototypes Layers to QBAFs}
\label{Super-Prototypes Layers to QBAFs}

% \todo[inline]{let's think a bit more about the 
% story in this section :)}
Since the SMLP mimics what an \mlp\s does, it can be converted to a QBAF, similar to the approach followed in 
SpArX \cite{AyoobiPT23},
 by first sparsifying the SMLP and then translating it to a QBAF
(c.f., Section~\ref{Preliminaries}).
% Using the similarity scores as input, \mlp\s is used for classification. After the training phase,  \mlp\s is converted to a QBAF
% (c.f., Section \nameref{pre} -- this involves sparsifying the underlying \mlp\s and then translating it to a QBAF). 
SpArX sparsifies an MLP by merging similar neurons.
Since we have  activation maps instead of single neurons,
we have to redefine the distance function in SpArX.
% Since activation maps are used in SMLPs instead of single neurons in MLPs, the definition of distance $\delta$ between two 
Given an input $x$ and two activation maps
$A^l_{i}$ and $A^l_{j}$ with height $H$ and width $W$,
% for a target input  is adapted.  For all the locations in the activation maps and for all neighbouring random samples $x' \in \Delta^\prime$ (as defined in SpArX) with sample weight $\pi_{x^\prime, x}$, 
our distance function is defined as:
\begin{align}
    \delta(A^l_{i}, A^l_{j}) =  \sum_{x'\in\Delta^\prime}\pi_{x', x}\sqrt{\sum_{h = 0}^{H-1}\sum_{w = 0}^{W-1}\big(A^l_{i} - A^l_{j}\big)^2_{h, w}},
\end{align}
where $\Delta^\prime$ denotes a sample neighborhood of $x$
and $\pi_{x', x}$ is a similarity function that assigns a
higher weight to neighbors $x'$ closer to the input $x$
\cite{AyoobiPT23}.

The obtained QBAF can explain how prototypes
reason for or against a particular class.
To illustrate 
% , making ProtoArgNet interpretable as we illustrate next. 
% \todo[inline]{Figure shows the provided reasons for and against assigning a specific bird to a specific class in the form of a QBAF. }
% \subsubsection{Sparsification of MLPs}
% \subsubsection{MLPs to QBAFs}
%\subsection{Illustrative Example}
% \label{example}
% \begin{example}
%For illustration, the 
the idea, consider the (sparsified) 1-hidden layer-SMLP/QBAF in Figure~\ref{fig:architecture}. It can be interpreted as follows:
\begin{itemize}

    \item The prototypes of the beak and eye support with high intensity the top-most hidden activation map, the prototypes corresponding to the neck and the upper wing support the middle hidden activation map, and the prototype of the tail supports the bottom hidden activation map%,  respectively;
    .
    \item The top and middle hidden activation maps (arguments) strongly support the super-prototype of the target class ``Baird Sparrow'' which forms the head, while the bottom hidden activation map attacks it. This leads to a super-prototype with positive values for the head (green overlay) and negative values for the tail (red overlay). %This leads to a similarity score of 0.7. 
    % This can be translated to ``The head of the input looks like a Baird Sparrow while the tail looks atypical (considering the observed training data)''. % corresponding to, respectively, the bottom and top clusters in the first hidden layer
    % ;
   %  \item 
   % The output neuron for class ``Baird Sparrow'' is strongly supported by the top and middle hidden clusters, but weakly attacked by the bottom hidden cluster. This leads to a high confidence level (with a probability of 0.9) for predicting the output class as "Baird Sparrow". 
    % the hidden clusters and output neurons are visualized using the super-prototypes they ``propagate'' through the SMLP, in the sense that these super-prototypes support them, e.g. the super-prototype for Class 0 supports the top cluster in the  hidden layer %and, indirectly, the top cluster in the second hidden, 
    % and the predicted Class 1 is supported by the super-prototype for Class 1.
\end{itemize}
%Note that the colours in the input images are irrelevant to the classification task here (which associates Class 1 to images with a triangle in the left column and a circle in the right column on the same row, no matter their colour), and the colours in the super-prototypes indicate support (green) for Class 1 at the bottom and attack (red) against Class 0 at the top. %\todo{massage inside this description ``The green colour in the super-prototypes shows the supported regions in the input image and the red colour shows the attacked/unsupported regions for assigning a certain class to the input''?} 
% (since we are dealing with binary classification here, the supporting regions for accepting one class are the attacking regions for accepting the other class). 
Overall, this interpretation indicates that
the predicted class for the input image is supported by the bird's head that looks like a ``Baird Sparrow'' and attacked by the tail differs from that,
while also pointing to the reasoning of the SMLP in terms of the prototypes used.
\section {Training ProtoArgNet}
\label{Training ProtoArgNet}

% \todo[inline]{This section is a bit vague. What is the motivation for the loss terms. How are they related to existing
% loss terms in the literature. For example, is cls related to
% the clustering loss? Explain what SP and cls stand for. The projection phase may require some clarification for readers 
% unfamiliar with the ProtoPNet idea}

% \todo[]{Be more specific about which training phases of which prototype approaches
% we do not need and why this is relevant? We still have the projection "phase"? H: Yes, the projection phase is for the prototypes. Since we have not used Gumble-Softmax trick. Instead, we use normal projection.
% }
ProtoArgNet is trained end-to-end
and does not require a prototype pruning
stage as some approaches do (e.g. ProtoPNet~\cite{chen2019looks}, ProtoPShare~\cite{RymarczykST021}, ProtoTrees~\cite{NautaBS21}, and PIP-Net~\cite{Nauta23}).
% This means that all amongst the prototype, and the super-prototype layers are trained at once without a need for additional pruning step. 
For the $i^{th}$ data point in a dataset of size $n$, with the data point belonging to class label $y_i \in K$ (where $K$ is the set of class labels), the target class super-prototype should obtain a high similarity score $\simscore^{y_i}$. Moreover, the corresponding similarity scores for the super-prototypes of other classes
% ($\{\simscore^{k}\}_{\substack{k=1, k\neq y_i}}^{|K|}$% in the dataset
% ) 
should be low. Simultaneously, the output of the classifier should be 1 for the target class $y_i$ and 0 for the other classes. These two objectives are aligned and can be implemented by a single loss function $L_{\superprototype}$. Additionally, we would like the prototypes learned by the model to be dissimilar to each other to encourage diversity by incorporating a dissimilarity loss $L_{\dis}$. 
% \begin{definition}
The \emph{total loss function} that we aim to minimize is: 
% \todo{$\loss$ was the number of hidden layers before....confusing. May be use Loss?}

\begin{align}
    %\min_{W_{conv}, W_{\prototype}, W_{\superprototype}, W_{\mlp\s}} \sum_{k=1}^{K}
    Loss =  L_{\superprototype} + \alpha L_{\dis}
\end{align}
where 
% $L_{\superprototype}$ is for the super-prototype (SP) kernel and $L_{cls}$ is for the classifier (cls):  
% \begin{itemize}
% \item 
$L_{\superprototype}$ 
% aims to penalize wrong predictions of the class labels%. Considering that the proposed ProtoArgNet model $M$ contains a convolutional backbone, a prototype layer, a channel-wise max layer, a super-prototypes layer, and a classification layer, the $L_{cls}$ is defined as below. 
% , using 
is the Cross-Entropy loss% (\textit{CrsEnt})  as follows, 
%\todo{what is CrsEnt? not sure...but may be it is standard?}
% \begin{align}
%     L_{CE} = \sum_{i=1}^{n} CrsEnt(G(x_i), y_i),
% \end{align}
% where $G(x_i)$ denotes the output of ProtoArgNet.
%     \item
, $\alpha$ is a constant ($\alpha = 0.1$ is used in the experiments) and
$L_{\dis}$ is defined as

\begin{align}
    L_{\dis} = \sum(|P.P^\intercal - I_{N}|),
\end{align}
% \end{itemize}
% \end{definition}
where $P$ is the matrix of all normalized prototypes, $I_{N}$ is the identity function of size $N \times N$, and $|.|$ is the absolute value function. Note that by definition of
$P.P^\intercal$, the entry at position
$(i,j)$ contains the dot product of the
$i$-th and $j$-th normalized prototypes.
All elements of the main diagonal are equal to 1. The non-diagonal elements correspond to the cosine similarities between pairs of prototypes and are $0$ if and only if the prototypes are
orthogonal. Hence, the loss term will be minimal
if all prototypes are orthogonal,  thus encouraging diversity among them.

We minimize our loss function
using the AdamW optimizer \cite{adamw}.
The trainable parameters are the convolutional weights $\convWeights$, prototypes $\prototype$, hidden layers weights $W^{l}$, and super-prototype weights $W^{\superprototype}$.
\begin{equation}
    \min_{\convWeights, \prototype, W^{l}, W^{\superprototype}} Loss(\convWeights, \prototype, W^{\l}, W^{\superprototype})%Do you mean like this?
\end{equation}
%
% \subsection{Projection of Prototypes}
After training, we perform a \emph{projection
step} analogous to ProtoPNet \cite{chen2019looks}. That is, we replace
each learnt prototype with the latent representation of the closest image patch from the training data. This allows
associating each latent prototype with
an image space representation %given by the image patch from the training data
(see the image patches in Figure \ref{fig:architecture} for an illustration).
% that each prototype has a global interpretable representation from the input space.  

\section{Experiments}
\label{Experiments}
% \todo[inline]{in the experiments, it may be good to look not only at 
% learning performance, but also at the interpretability of the
% identified prototypes. A general comparison may be difficult 
% (perhaps we can still brainstorm a little bit about it),
% but we could probably add some examples to give a direct 
% comparison between the previous prototypes and ours and discuss
% advantages/disadvantages}

% \todo[inline]{Li et al have a nice ablation study in Section 4 that
% shows how they make sure that prototypes are meaningful. Perhaps we
% can do something similar?}

% \todo[inline]{Hamed: The structure of the Experiments section. 1) Ablation study 2) Accuracy vs. Interpretability: We can do high-level representation of super-prototypes, Check Dataset CelebA 3) Spatial Correlation between objects. In which cases it can be useful? How useful it is in other cases? Is it even worse for the other cases? If not why? If yes why?}

We compared ProtoArgNet to the state-of-the-art prototypical-part-learning models ProtoPNet \cite{chen2019looks}, ProtoTrees \cite{NautaBS21}, ProtoPShare \cite{RymarczykST021}, ProtoPool \cite{protoPool} and PIP-Net \cite{Nauta23}. Our experiments (set-up in Section~\ref{sec:set-up}) evaluate the classification %scores
performance (Section~\ref{Classification}), the sparsification process (Section~\ref{sec:sparse}), the role of each layer on the model's %prediction accuracy 
performance by an ablation study (Section~\ref{Ablation}), the ability to encode and detect spatial relationships in the input (Section~\ref{sec:Spatial Correlation}) and the complexity of explanations drawn from ProtoArgNet (Section~\ref{Complexity}).
% and the cognitive tractability of the local explanations (Section~\ref{Cognitive Tractablity})%\todo{add cognitive tractability here as well? Is the meaning of "cognitive complexity" and "cognitive tractability" clear? See also later comment at the end of tractability section}
As usual, we use top-1 accuracy (the standard accuracy) as the performance measure.
% , as is the case with the baselines.
We also perform a qualitative evaluation (Section~\ref{sec:qual}).

For all %the first three 
experiments, we have used CUB \cite{CUB_200_2011} and Cars \cite{CARS}, which are the standard benchmarks for prototypical-part-learning models.\footnote{
For the ablation study, we include additional experiments with small-scale datasets (MNIST% Digits
~\cite{mnist}, Fashion MNIST~\cite{fashionMNIST}, CIFAR10~\cite{cifar10} and GTSRB \cite{gtsrb}) in the supplementary material.}
% \todo[inline]{Remove the small-scale datasets from Table 2 and include them as additional experiments for SM.}
% , and the CUB and Cars large-scale datasets.
To emphasize the importance of localizing specific regions in images that either support or attack the target class, we utilized the Brain Tumor MRI dataset \cite{brainMRIdataset}.
Additionally, we demonstrate ProtoArgNet's capability to identify spatial relationships that are undetectable by other approaches by applying it to (an adaptation to binary classification of) the SHAPES  dataset \cite{SHAPES}.  

\subsection{Experimental Setup}
\label{sec:set-up}
% \todo[inline]{all choices in this subsection need to be justified}
% \todo[inline]{I suppose that this is the standard procedure (resizing)? Then we should say that, otherwise it may raise questions ;)}
Following the usual protocol \cite{chen2019looks}, the input images are resized to $224 \times 224$. We set the number of prototypes $N$ to 1024.\footnote{The performance of ProtoArgNet with various choices for $N$ (512, 1024, 2048) is reported in the supplementary material.} For training the model, we set the batch size to 32 and the number of epochs to 1000. %\todo{why these numbers? have we tried others? are details in the supplementary material?} 
The convolutional backbone was ResNet-50~\cite{resnet} pre-trained using ImageNet~\cite{imageNet}. The choices of batch size, number of training epochs, convolutional backbone and pre-trained weights are aligned with previous prototypical-part-learning approaches. %\todo{why the choice of ResNet? why trained on ImageNet?} 
\setlength{\tabcolsep}{2.3pt}
\begin{table}
\centering
\vspace{-3mm}
\begin{tabular}{c c c c|c}
\hline
\multirow{2}{*}{Method} & \multicolumn{3}{c}{Accuracy} \\
\cline{2-5}
 & \multicolumn{1}{c}{CUB} & \multicolumn{1}{c}{Cars} & \multicolumn{1}{c|}{Brain} &\multicolumn{1}{c}{SHAPES} \\
\hline
ProtoPNet & 79.2 \size{8}{$\pm$} 0.1 & 86.1 \size{8}{$\pm$} 0.1 & 97.4 \size{8}{$\pm$} 0.2& 50.6 \size{8}{$\pm$} 0.7  \\
ProtoPShare & 74.7 \size{8}{$\pm$} 0.2 & 86.4 \size{8}{$\pm$} 0.2& 97.7 \size{8}{$\pm$} 0.1 & 50.2 \size{8}{$\pm$} 0.8 \\
ProtoPool & 80.3 \size{8}{$\pm$} 0.2  &  88.9 \size{8}{$\pm$} 0.1& 98.3 \size{8}{$\pm$} 0.2 & 49.7 \size{8}{$\pm$} 0.6 \\
ProtoTrees & 82.2 \size{8}{$\pm$} 0.7 & 86.6 \size{8}{$\pm$} 0.2& 98.0 \size{8}{$\pm$} 0.3 & 50.1 \size{8}{$\pm$} 0.7 \\
PIP-Net & 82.0 \size{8}{$\pm$} 0.3 & 86.5 \size{8}{$\pm$} 0.3  & 97.5 \size{8}{$\pm$} 0.3 &50.3 \size{8}{$\pm$} 0.6 \\
ProtoArgNet & \textbf{85.4 \size{8}{$\pm$} 0.2} & \textbf{89.3 \size{8}{$\pm$} 0.3} & \textbf{99.5 \size{8}{$\pm$} 0.3}&  \textbf{99.8 \size{8}{$\pm$} 0.1} \\
\hline
\end{tabular}
\caption{Accuracy of ProtoArgNet and other prototypical-part-learning methods on the CUB, Cars, Brain and SHAPES datasets.  SHAPES is used for the evaluation of spatial correlation between prototypical-parts% in images
. (Best %results 
    accuracy in {\bf bold})}
\label{tab:classification}
\end{table}
The SMLP %\todo{perhaps introduce and use another term than MLP? Perhaps MMLP for matrix-MLP?}
had 1 hidden layer,
 400 hidden activation maps and GELU activation functions\footnote{The performance of ProtoArgNet employing various MLP configurations, encompassing 1 to 5 hidden layers and a range of hidden activation maps (50, 100, 200, 400, 600), is detailed in the supplementary material.}. 

% \todo{this must be clarified. Of course, sparsification will at some point negatively affect performance. Shouldn't we simply say how we parametrized the sparsification?}
%\todo{why these choices?} \todo{why then two hidden layers in Fig 2? I probably misread it, and the first hidden layer is the input? so that super-prototypes are external? my illustrative example need s fixing...but then why Figure 4 has one less layer?}

% \todo[inline]{Change geometrical to spatial everywhere.}

\subsection{Classification %Scores
Performance}\label{Classification}
The first two columns in Table \ref{tab:classification} show the accuracy of our method compared to %other prototypical-part-learning approaches ProtoPNet, ProtoPShare, ProtoPool, ProtoTrees, and PIP-Net
the CUB, Cars and Brain datasets. For all datasets, our ProtoArgNet outperforms the baselines.% with {83.4 $\pm$ 0.2} and {89.3 $\pm$ 0.2} top-1 accuracies, respectively
% ProtoArgNet is the only approach that 

\subsection{Sparsification of QBAF}
\label{sec:sparse}

To evaluate the tradeoff between sparsity and performance, 
we evaluated the accuracy under $40\%,  80\%, \text{and } 94\%$ sparsification
ratios (240, 80, and 24 activation maps remaining).
As Table \ref{tab: sparsification} shows, the classification accuracy
remained unchanged up to the $93\%$ ratio. At $94\%$ sparsification, the accuracy starts dropping. However, we can see that the size of the SMLP 
can often be reduced significantly without affecting its performance 
negatively. This is in line with the experiments in~\cite{AyoobiPT23}. 
% Using the SpArx methodology, we were able to sparsify the QBAFs
% by 80\% (from 400 to 80 activation maps) without compromising classification accuracy as shown in . 
% This is in line with SpArX~\cite{AyoobiPT23}.

\setlength{\intextsep}{0.5pt}
\begin{table}[]
\centering
\begin{tabular}{c c c c c}
\hline
Sparsification&\multicolumn{4}{c}{Datasets}\\
\cline{2-5}

 Ratio& CUB & Cars & Brain & SHAPE\\
\hline
% 0.2  & 85.4 & 89.3 & 99.5 & 99.8 \\
0.4 & 85.4 & 89.3 & 99.5 & 99.8 \\
% 0.6 & 85.4& 89.3 & 99.5 & 99.8\\
0.8 & 85.4 & 89.3 & 99.5 & 99.8 \\
0.94 & 85.1 & 88.8 & 99.4 & 99.8 \\
\hline
\end{tabular}
\caption{Classification accuracy of ProtoArgNet with different sparsification ratios for different datasets. As in SpArX \cite{AyoobiPT23}, the local explanations with various sparsification ratios have not compromised classification accuracy.}
\label{tab: sparsification}
\end{table}

\subsection{Ablation study}\label{Ablation}
Ablation studies on CUB and Cars (excluding the Brain and SHAPES datasets, which have not been used for the baseline methods) in Table \ref{tab:ablation} show that ProtoArgNet achieves the best accuracy when employing a cosine similarity prototype layer with an SMLP with one hidden layer. 
% Alternatively, the L2-distance-based prototype-layer, as utilized in ProtoPNet, can be employed in conjunction with a fixed logistic regression layer for classification (fine-tuned in the second training phase in ProtoPNet). 
Notably, ProtoArgNet surpasses the performance of state-of-the-art methods even when utilizing a fixed logistic regression layer, instead of SMLP super-prototype layers (but performs best with the SMLP).

\setlength{\intextsep}{0.5pt}%
\setlength{\tabcolsep}{2.3pt}
\begin{table}
    \centering
% \fontsize{9}{9}\selectfont
\begin{tabular}{ccccc} \hline
      Super- & Prototype & \multirow{2}{*}{Classifier } & \multicolumn{2}{c}{Accuracy} \\ \cline{4-5} Prototypes& Layer &
      &  CUB & Cars \\ \hline
    \longdash[3] &L2 & Fixed  &  79.5 & 86.4 \\
     \longdash[3] &L2 %(Our) 
     &  SMLP & 81.5 & 86.9
       \\ %\midrule
     \longdash[3] &Cosine & Fixed  & 81.7 & 87.4  \\
     \longdash[3] &Cosine 
     & SMLP  & 81.9 & 88.0
     \\ 
         \checkmark &L2 & Fixed  &  81.4 & 87.6 \\
     \checkmark &L2 %(Our) 
     &  SMLP & 82.7 & 88.3
       \\ %\midrule
     \checkmark &Cosine & Fixed  & 83.5 & 88.9  \\
     \checkmark &Cosine 
     & SMLP  & \textbf{85.4} & \textbf{89.3}
     \\\hline
\end{tabular}
    \caption{Ablation study with different prototype layers 
    and classifiers with %the presence of 
    respect to the super-prototypes.}
    \label{tab:ablation}
\end{table}

% \todo[inline]{Hamed: SparX with different compression ratios. }
% \subsection{Interpretability}
% \todo[inline]{In this section, we just visualize the prototypes for ProtoArgNet together with the QBAF and other approaches and compare the representations}

\setlength{\intextsep}{5pt}%
\begin{figure}
    \centering
    \vspace{-3mm}
    \includegraphics[width=1\linewidth]{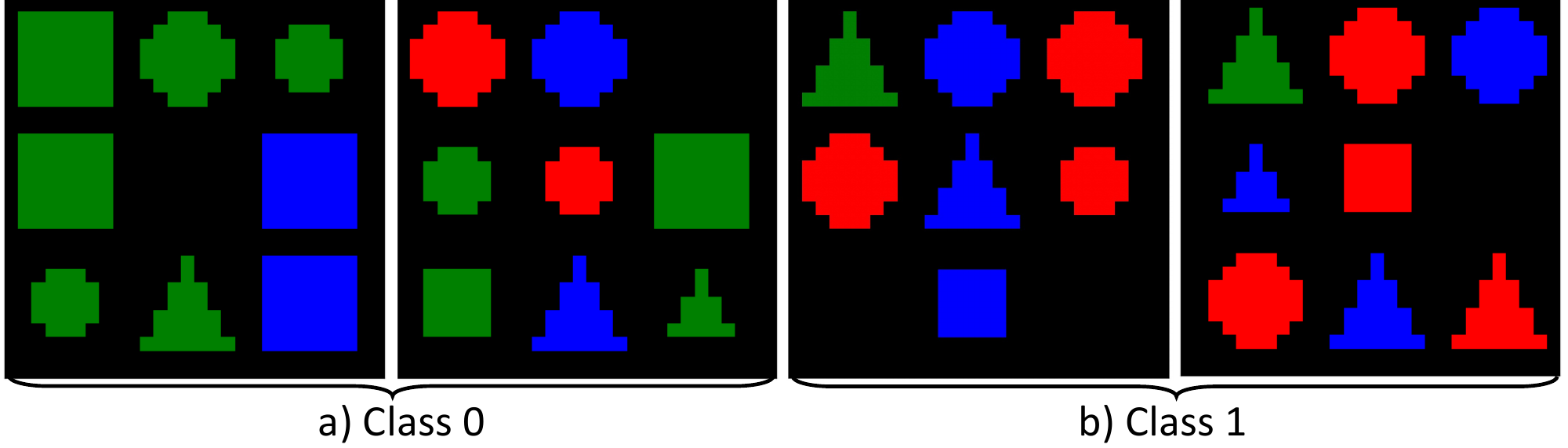}
    \caption{Examples from (our binary adaptation of) the SHAPES dataset. Class 1 contains images with a triangle in the leftmost column and a circle in the rightmost column of the same row or vice versa. Class 0 is when these conditions are not met.}
    \label{fig:SHAPES}
\end{figure}

\begin{figure*}[hbt!]
    \centering
    \vspace{5mm}
    \includegraphics[width=1\linewidth]{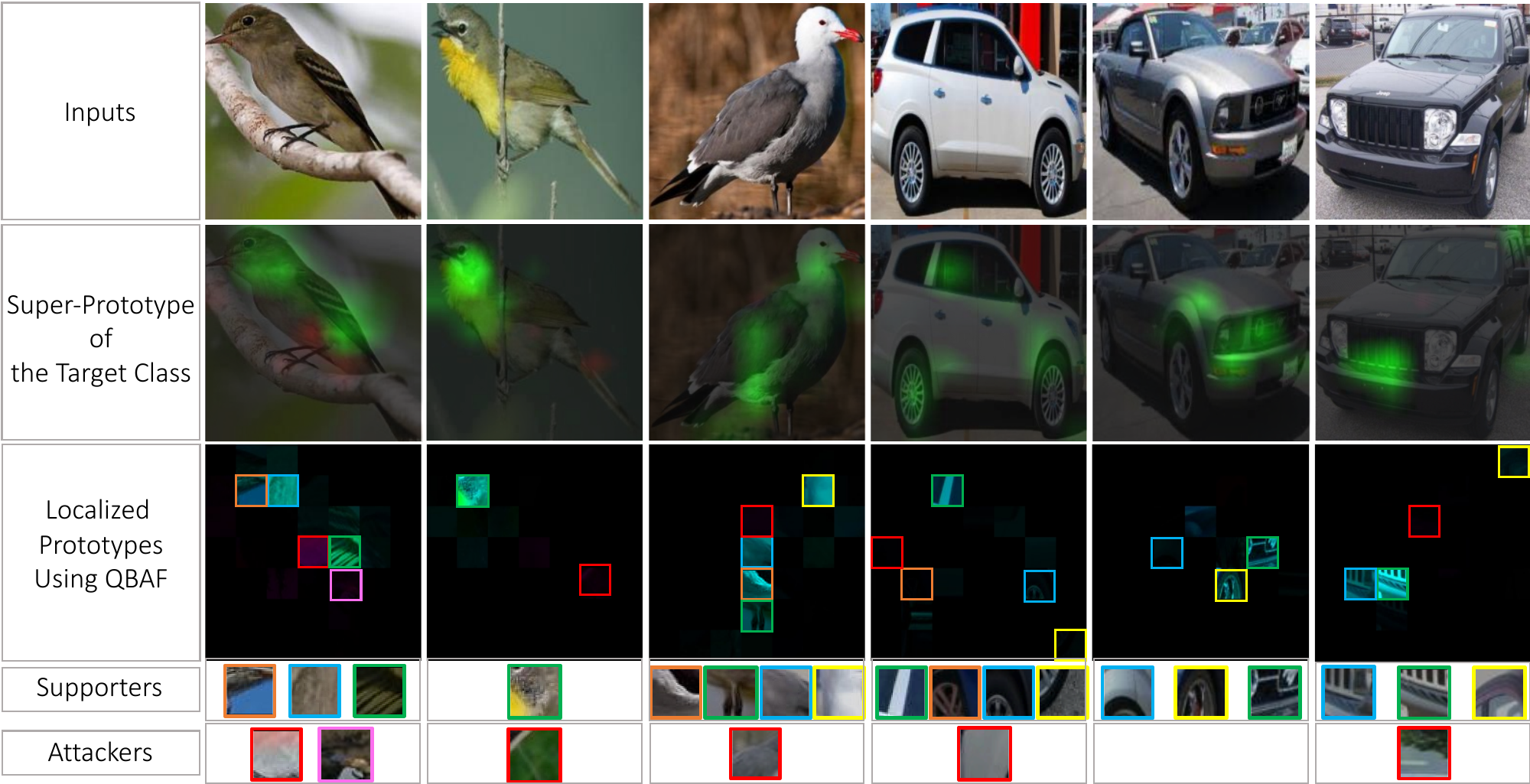}
    \caption{Examples of ProtoArgNet explanations for the CUB and Cars dataset. The first row shows input images. The second row shows the super-prototypes of the target classes provided to the user as explanations. The third row shows the corresponding localized activated prototypical-parts for each 
 super-prototype visualized by following the attack and support relations in the QBAF. The last two rows show the corresponding supporting prototypical-parts and attacking prototypical-parts. 
    % aciprototypical-parts are learned from the image patches in the training set. The super-prototype highlights the supported regions with a green overlay and the attacked / unsupported regions with a red overlay. The output neuron in the MLP/QBAF assigns the probability of 0.9 for classifying the input image as a \emph{Baird Sparrow}.
    % The QBAF outlines the reasoning of the MLP.
    }
    \label{fig:super-prototypes}
\end{figure*}
\subsection{Localization and Spatial Correlations}\label{sec:Spatial Correlation}
Figure \ref{fig:brain} shows some super-prototypes for randomly selected examples from the Brain Tumor MRI dataset. These images showcase the regions that are either supporting or attacking the corresponding classes. For instance, when examining the leftmost input image, a radiologist
would find that the `Meningioma' (benign tumour) class has the highest probability. She can then look into the plausibility of
the decision process by looking at
the corresponding super-prototype (leftmost super-prototype). The prototypical-parts associated with greenly highlighted regions should
be indicative of `Meningioma' (benign tumour), while those in the redly highlighted regions should be contraindicative. To understand the significance of the red-highlighted area, the radiologist can compare it with the super-prototypes of other classes where the same regions are highlighted in green. The second super-prototype from the left, associated with the Glioma (malignant tumour) class, highlights these regions in green, suggesting that further examination of that region may be necessary. 
% Looking at the green regions (showing the supporting prototypes), they can locate the probable tumour \todo[inline]{continue...}
% for the target class `Meningioma' (benign tumour),
% radiologists might interpret the super-prototype as indicating a high probability and could locate the supporting region with high confidence (indicated by the intense green). However, the presence of a light red area (indicating a low confidence) in the image might indicate the presence of Glioma (malignant tumour), suggesting that further examination of that region may be necessary. 

To assess whether different image classification methods can localize the prototypical-parts and encode spatial relationships between them, we adapted the SHAPES dataset \cite{SHAPES} as a benchmark. 
We randomly generated synthetic images containing $3 \times 3$ grids of circles, triangles, and squares in different colours (red, green, and blue).  An image is assigned to Class 1 if a triangle is located in the leftmost column and a circle is located in the rightmost column of the same row or vice versa, with a circle in the leftmost column and a triangle in the rightmost column of the same row\footnote{This criterion can be customized to reflect the user's preferences, e.g.  the dataset could assign disjunction of multiple criteria.}, and Class 0 otherwise. The resulting dataset comprises 10,000 $ 224\times224$ images with balanced binary class labels.
Figure \ref{fig:SHAPES} shows examples of images in the dataset. 
%The green colour in the super-prototypes shows the supported regions in the input image and the red colour shows the attacked/unsupported regions for assigning a certain class to the input.     

% \begin{figure*}[h]
%     \centering
%     \includegraphics[width=1\linewidth]{figures/SHAPES_SP.pdf}
%     \caption{ProtoArgNet encoding the spatial correlation between different prototypical-parts in the class-wise super-prototypes. The green colour in the super-prototypes shows the supported regions in the input image for choosing a specific class label and the red colour shows the attacked regions (reasons against choosing a specific class).  }
%     \label{fig:SHAPES}
% \end{figure*}

% \begin{table*}[t]
% \centering
% \caption{}
% \label{tab:spatial}
% \begin{tabular}{|c|c|c|}
% \hline
% \multicolumn{1}{|c|} {} & \multicolumn{1}{|c|}{SHAPES}  \\
% \hline
% Method & Top-1 accuracy  \\
% \hline
% ProtoPNet & 51.1 $\pm$ 0.7  \\
% ProtoPShare & 50.4 $\pm$ 0.8  \\
% ProtoPool & 50.8 $\pm$ 0.6   \\
% ProtoTree & 51.4 $\pm$ 0.7   \\
% PIP-Net & 50.6 $\pm$ 0.6  \\
% ProtoArgNet (our) & \textbf{98.4 $\pm$ 0.2} \\
% \hline
% \end{tabular}
% \end{table*}

The last column in Table \ref{tab:classification} compares the %top-1 %classification 
accuracy of %different prototypical-part-learning approaches 
the baselines for this SHAPES dataset. ProtoArgNet, with an accuracy of 99.7\% ± 0.2\%, significantly outperforms all other approaches (whose accuracy  is around 50\%). This can be explained by noting that these models only 
look at the presence of prototypes in images, %and require additional HERE
but are unable to infer information 
from their relative position. ProtoArgNet addresses this limitation by using channel-wise max and super-prototypes, which enable the model to infer the spatial correlation of different prototypical-parts in the image when needed for classification.

% \todo[inline]{In this section, we compare all the methods for the classification of the SHAPES dataset with different queries like a red object is on the left side of a green object. This might have applications in autonomous vehicles as well when one needs to detect the exact location of objects or in image segmentation. Figure \ref{fig:SHAPES-SP} shows an example of a learned super-prototype by emphasizing the 

% 1. A table to report top-1 accuracies for SHAPES

% 2. Visualization of SHAPES outputs for ProtoArgNet}

\subsection{Cognitive Complexity of Explanations}
\label{Complexity}

% The super-prototypes can serve as the basis for human-readable explanations for the outputs of ProtoArgNet. 
% Figure \ref{fig:architecture} showed a generated local explanation for a data point in SHAPES (see the examples in Section~\nameref{ProtoArgNet} for details on this illustration). 
%

% We can use the number of representative \text{prototypes} to measure the cognitive complexity of the explanations drawn from prototypical-part-learning methods. Since ProtoArgNet maintains the $7\times7$ dimensions of the convolutional output of the ResNet50 backbone and employs the maximum value across the number of \cwms\s during inference, thereby ensuring that, at most, a single prototype is activated at each location. In theory, the upper bound of the number of activated prototypes for ProtoArgNet is $7\times7=49$ per example. For the other approaches, according to Table \ref{tab: num_proto}, the upper bound of the activated prototypes per example is the total number of prototypes: 2000, 495, 400 and 202 for the CUB dataset and 2000, 515, 480 and 195 for the Cars dataset for the ProtoPNet, PIP-Net, ProtoPshare, and ProtoTrees, respectively.
% The cognitive complexity of explanations generated by prototypical-part-learning methods can be measured by the number of representative activated prototypes. 
We measure the \emph{cognitive complexity} of an explanation of a prototypical-part-learning approach by the number of activated prototypes per input $x_i$. We consider a prototype $p_j$ to be activated if the maximum value in its similarity map (in $\simmap_j$) exceeds a threshold of $\tau=0.1$ (after normalizing the absolute value of the similarity scores to the range [0, 1]).

ProtoArgNet retains the $7\times7$ spatial dimensions of the convolutional output from the ResNet50 backbone, applying maximum function across \cwms\s during inference. This ensures that at most a single prototype is activated per spatial location for each input. Consequently, the theoretical upper bound for activated prototypes in ProtoArgNet is $7\times7\!=\!49$ per example. In contrast, the upper bound for %the number of
activated prototypes in other methods equals the total number of prototypes (see  Table \ref{tab: cognitive_tractability})
: 2000, 495, 400, and 202 for the CUB dataset, and 2000, 515, 480, and 195 for the Cars dataset for ProtoPNet, PIP-Net, ProtoPshare, and ProtoTrees, respectively. Notably, ProtoArgNet uses these 49 prototypes to construct the super-prototypes of all classes. Further, Table \ref{tab: cognitive_tractability} reports the Average number of Activated Prototypes per example ($\#AAP$),  confirming that ProtoArgNet has lower cognitive complexity %compared to 
than other methods.

\subsection{Qualitative Evaluations}
\label{sec:qual}

  The super-prototypes of the target classes in the bottom row of Figure \ref{fig:super-prototypes} illustrate the local explanations generated for a few data instances from the CUB and the Cars datasets. The top row shows the input images to ProtoArgNet. To interpret the super-prototypes, one could trace the attack and support relations in the QBAF to localize the prototypical-parts on the super-prototypes as in the third row.  
  % specifically for the target class ``Baird Sparrow." 
  The green overlay on the super-prototypes highlights the regions in the input image that support the classification, while the red areas identify the attacked or unsupported portion of the input.
  For example, the super-prototype of the left-most image in Figure \ref{fig:super-prototypes} can be interpreted as the bird's neck and the wing resembling the target class while the belly and flank differ from the observed target class instances in the training set.  
%   This super-prototype can be 
% interpreted as the bird's head resembles a ``Baird Sparrow," but its tail is atypical for this species. We have added this reading manually here for illustration, simulating how a human may read the super-prototypes. 
We leave the automatic generation of human-readable interpretations of the super-prototypes and the QBAF for explanatory purposes to future work. 
% \todo{I am not sure if I understand the idea of complexity vs tractability (given their meaning in the theory of computation, I am also not sure if this combination of terms is a good one). Does complexity refer to the "globally" and tractability to the "locally" relevant prototypes? HA: Yes, cognitive complexity refers to the total number of prototypes/super-prototypes in the model that is intended for the explanation which is a global measure and cognitive tractability refers to the average number of activated prototypes per example which is a local measure.  }
\setlength{\intextsep}{0.5pt}
\begin{table}[]
\centering
\begin{tabular}{c c c }
\hline
\multirow{2}{*}{Method}&\multicolumn{2}{c}{\# $AAP$ / Upper Bound }\\
\cline{2-3}

&CUB & Cars\\
\hline
ProtoPNet & 1147.84/2000 & 1059.49/2000   \\
ProtoPShare & 182.58/400 & 159.20/480  \\
ProtoPool & 95.19/202& 70.64/195  \\
ProtoTrees & 103.78/202 & 76.31/195 \\
PIP-Net & 211.83/495 & 213.46/515 \\
ProtoArgNet & \textbf{24.57/49} & \textbf{8.42/49}  \\
\hline
\end{tabular}
\caption{Comparing the Average number of Activated Prototypes ($\# AAP$) per example and the upper bound of activated prototypes for each method. A lower number indicates lower cognitive complexity.}
\label{tab: cognitive_tractability}
\end{table}

\section{Conclusion}
ProtoArgNet is a novel prototypical-part-learning approach.
It utilizes super-prototypes that combine multiple 
prototypical-parts to a single class representation that
can take account of spatial relationships
between individual parts.
% ProtoArgNet can be trained end-to-end and does not require 
% an additional pruning phase of prototypes. 
% As opposed to previous 
% prototypical-part-learning approaches, the use of super-prototypes
% allows ProtoArgNet to capture spatial relationships between prototypical-parts. 
% although convolutional neural networks (CNNs) can encode these correlations. 
% ProtoArgNet addresses this limitation with the super-prototypes layer. 
Using an MLP structure for the super-prototypes layers allows ProtoArgNet to capture
non-linear relationships, while
applying the SpArX methodology allows interpretable argumentative reading of the MLP as a QBAF.
% \begin{table}
% \centering
% \vspace{4mm}
% \begin{tabular}{ c c c}
% \hline
%  Compression & \multicolumn{2}{c}{Accuracy}  \\
% \cline{2-3}
%  Ratio & CUB & Cars \\
% \hline
%  0.2 &83.1 $\pm$ 0.2 & 89.2 $\pm$ 0.2   \\
%  0.4 &82.9 $\pm$ 0.1 & 88.9 $\pm$ 0.1  \\
%  0.6 & 82.5 $\pm$ 0.2  &  88.4 $\pm$ 0.1 \\
%  0.8 &81.8 $\pm$ 0.2 & 88.1 $\pm$ 0.2  \\
% \hline
% \end{tabular}
% \caption{Accuracy of ProtoArgNet with different compression ratios of the MLP for the CUB and Cars datasets.}
% \label{tab: MLP}
% \end{table}
%%%%%%%%%%%%%%%%%%
% By ProtoArgNet uses multi-layer perception (MLP) instead of logistic regression for the classification layer. ProtoArgNet preserves the model's interpretability by converting the MLPs to an argumentation framework.
%
Experiments show that ProtoArgNet outperforms state-of-the-art prototypical-part-learning approaches in terms of accuracy, cognitive complexity, and the ability to learn spatial relationships between prototypical-parts. 

Future work includes expanding ProtoArgNet's capabilities further to encompass multi-modal data. 
% Additionally, we will investigate the implementation of a user-model feedback loop to enhance the debugging process for super-prototypes. Further, we plan to deploy ProtoArgNet in the medical domain. Finally, we plan to explore various options for obtaining explanations from ProtoArgNet, including interactive forms thereof~\cite{argXAIsurvey}.

% \todo[inline]{can we add future work? User experiments? Other unstructured data? multi-modal data? }

% \clearpage  % TODO REVIEW/FINAL: This \clearpage needs to be removed from both review and camera-ready versions.


\section{Acknowledgments}
This research was partially funded by the  European Research Council (ERC) under the European Union’s Horizon 2020 research and innovation programme (grant
agreement No. 101020934, ADIX) and by J.P. Morgan and by the Royal
Academy of Engineering under the Research Chairs and Senior Research
Fellowships scheme.  Any views or opinions expressed herein are solely those of the authors.

\clearpage
\setcounter{page}{1}
\maketitlesupplementary
\setcounter{table}{4}

\appendix
\setcounter{figure}{5}
\setcounter{table}{4}

\section{Prototypes Configurations}
The performance of ProtoArgNet with various choices for the number of prototypes $N$ (512, 1024, 2048) is reported in Table \ref{tab:Proto}. The results show that $N=1024$ is the best choice for CUB, Cars, and Brain datasets.

\section{SMLP/QBAF Configurations}
Table \ref{tab:MLP} shows the performance of ProtoArgNet employing various SMLP configurations, encompassing 1 to 5 hidden layers and a range of hidden activation maps (50, 100, 200, 400, and 600). The results show that an SMLP with 1 hidden layer and 400 activation maps achieves the best accuracy. All configurations are trained for 1000 epochs.
% \todo[inline]{Report the MLP configuration accuracies.}

\section{Additional Ablation Studies}
\label{sec:add_abl}
We have conducted further ablation studies with %additional 
popular small-scale datasets, including MNIST~\cite{mnist}, Fashion MNIST~\cite{fashionMNIST}, CIFAR10~\cite{cifar10} and GTSRB \cite{gtsrb} as shown in Table \ref{tab:add_abl}. The results suggest that cosine similarity outperforms L2-distance and SMLP is a better classifier than a logistic regression with initial fixed weights for the first phase of training. Moreover, the super-prototypes layer boosts the accuracy of the model in all cases. This confirms the results in the main body of the paper.

\section{More Qualitative Examples}
This section has visualised more qualitative examples for the CUB, Cars, and Brain datasets. 
The top row shows the randomly selected inputs, the middle row shows the super-prototypes of the target classes for the corresponding inputs, and the bottom row shows the localized prototypes following the attack and support relation in the QBAF on the resulting super-prototype. Notably, the spatial dimensions of the convolutional outputs vary across datasets: for the CUB and Cars datasets, the output is 7x7, whereas for the Brain dataset, it is 14x14. This difference is due to following the same approach as PIP-NET~\cite{Nauta23}, where a max-pooling layer is removed from the ResNet50 backbone to obtain more fine-grained prototypes.

\begin{table}
    \centering
% \fontsize{9}{9}\selectfont
\begin{tabular}{cccc} \toprule
      \multirow{2}{*}{$N$}  & \multicolumn{2}{c}{Datasets} \\ \cmidrule{2-4}  & 
       {CUB} & {Cars} & {Brain} \\ \midrule
    512    & 85.0$\pm$0.2 & 88.8$\pm$0.3 & 98.9$\pm$0.2  \\
     % 512 &32 %(Our) 
     %  & 82.9 & 88.1 &   
     %   \\ %\midrule
     % 512 &64   & 82.6 & 87.7 &   \\
     % 512 &128
     %   & 82.2 & 87.1  &
     % \\ 
     1024  
       & \textbf{85.4$\pm$0.2} & \textbf{89.3$\pm$0.3}  &\textbf{99.5$\pm$0.3}
     \\
     % 1024 &32 
     %   & \textbf{83.4} & \textbf{89.3}  &
     % \\
     % 1024 &64
     %   & 82.8 & 88.4 &
     % \\
     % 1024 &128 
     %   & 82.3 & 87.6 &
         % \\
     2048  
       & 85.2$\pm$0.3 & 89.0$\pm$0.2& 99.2$\pm$0.3
     %        \\
     % 2048 &32 
     %   & 82.9 & 88.6 &
     %        \\
     % 2048 &64 
     %   & 82.5 & 88.1 &
     %        \\
     % 2048 &128 
     %   & 82.1 & 87.8 &
     \\\bottomrule
\end{tabular}
    \caption{The accuracy of ProtoArgNet with different numbers of prototypes $N$. (Best %results 
    accuracy in {\bf bold})}
    \label{tab:Proto}
\end{table}
\begin{table}
    \centering
% \fontsize{9}{9}\selectfont
\begin{tabular}{ccccc} \toprule
      \multirow{1}{*}{\# Hidden} & \multirow{1}{*}{\# Hidden} & \multicolumn{3}{c}{Datasets} \\ \cmidrule{3-5} Layers & Activation Maps
      & {CUB} & {Cars} &{Brain} \\ \midrule
    1 &50   & 83.5 &  87.7& 97.6 \\
     1 &100 %(Our) 
      & 84.6 &  88.2  & 98.8
       \\ %\midrule
     1 &200   & 85.1 &  89.0 & 99.3  \\
     1 &400
       & \textbf{85.4} &   \textbf{89.3} & \textbf{99.5}
     \\ 
          1 &600
       & 85.0 &   88.7 & 98.9
     \\ 
     2 &50 
       & 83.1 & 87.2 & 97.3
     \\
     2 &100 
       & 84.3 & 87.8 & 98.5
     \\
     2 &200
       & 84.6 & 88.4 & 98.8
     \\
     2 &400 
       & 84.8 & 88.6 & 99.0
         \\
             2 &600
       & 84.2 &   88.1 & 98.5
     \\ 
     3 &50 
       & 83.8& 87.2& 98.1
            \\
     3 &100 
       & 83.5& 86.8& 97.4
            \\
     3 &200 
       & 82.9& 86.2 & 97.1
            \\
     3 &400 
       &82.5 & 86.0 & 96.6
    \\
                 3 &600
       & 82.4 &   85.5 & 96.3
     \\
     4 &50 
       & 82.8& 85.9 & 97.1
            \\
     4 &100 
       & 82.5 & 85.5 & 97.0
            \\
     4 &200 
       & 82.5 & 85.1 & 97.0
            \\
     4 &400 
       & 82.2& 84.5 & 96.3
    \\
                 4 &600
       & 82.1 &   84.3 & 96.1
     \\
     5 &50 
       & 81.8& 84.7 & 96.3
            \\
     5 &100 
       & 81.2& 84.5 & 96.0
            \\
     5 &200 
       & 80.6& 84.1 & 95.6
            \\
     5 &400 
       & 80.3& 83.9 & 95.3
     \\
                  5 &600
       & 79.8 &   83.4 & 94.2
     \\
     \bottomrule
\end{tabular}
    \caption{ProtoArgNet accuracy with different SMLP/QBAF configurations. (Best %results 
    accuracy in {\bf bold})}
    \label{tab:MLP}
\end{table}
\begin{figure*}[th!]
    \centering
    \includegraphics[width=1\linewidth]{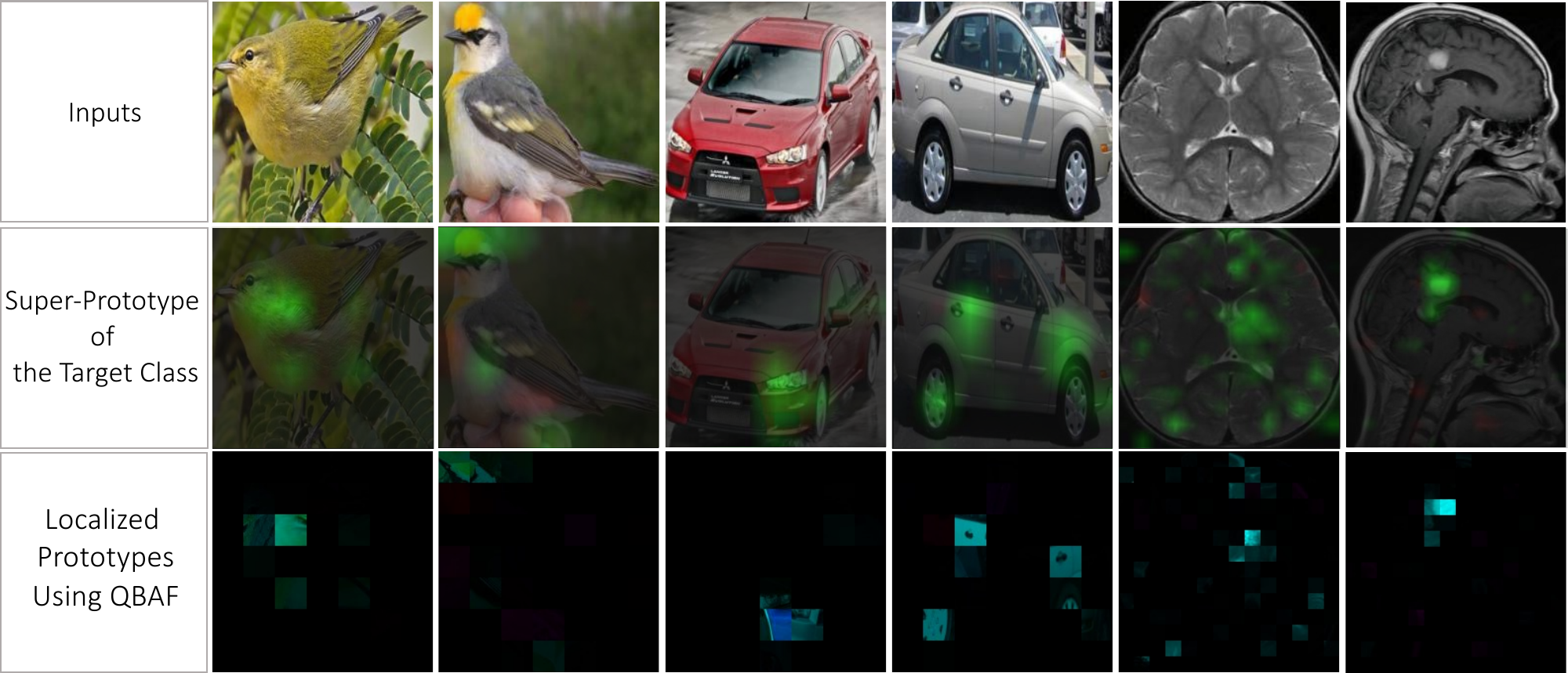}
    \caption{Examples of ProtoArgNet explanations for the CUB, Cars, and Brain dataset. The first row shows input images. The second row shows the super-prototypes of the target classes provided to the user as explanations. The third row shows the corresponding localized activated prototypical-parts for each 
 super-prototype visualized by following the attack and support relations in the QBAF. }
    \label{fig:CAR_EXPLANATION}
\end{figure*}
\begin{table*}
    \centering
% \fontsize{9}{9}\selectfont
\begin{tabular}{ccccccc} \toprule
      \multirow{2}{*}{Super-Prototype} & \multirow{2}{*}{Prototype} & \multirow{2}{*}{Classifier } & \multicolumn{4}{c}{Datasets} \\ \cmidrule{4-7} & &
      & {MNIST} & {Fashion} & {CIFAR10} & GTSRB  \\ \midrule
    \longdash[3] &L2 & Fixed  & 96.43 & 86.34 & 83.45 & 98.20  \\
     \longdash[3] &L2 %(Our) 
     &  SMLP & 97.39 & 87.53  & 83.98 & 98.37 
       \\ %\midrule
     \longdash[3] &Cosine & Fixed  & 97.98 & 88.21 & 84.22 & 98.78   \\
     \longdash[3] &Cosine
     & SMLP  & 98.30 & 88.75  & 84.51  & 99.36 
     \\ 
     \checkmark &L2 
     & Fixed  & 97.52 & 87.93  & 83.94  & 98.48 
     \\
     \checkmark &L2 
     & SMLP  & 97.75 & 88.39  & 84.57  & 99.33
     \\
     \checkmark &Cosine
     & Fixed  & 98.82 & 89.76  & 84.93  & 99.45
     \\
     \checkmark &Cosine 
     & SMLP  & \textbf{99.20} & \textbf{90.43}  & \textbf{85.31}  & \textbf{99.83} 
     \\\bottomrule
\end{tabular}
    \caption{Ablation study with different prototype %layers 
    and classifier layers with %the presence of 
    respect to a super-prototype layer for additional small-scale datasets. (Best %results 
    accuracy in {\bf bold})}
    \label{tab:add_abl}
\end{table*}

{
    \small
    \bibliographystyle{ieeenat_fullname}
    \bibliography{main}

\begin{thebibliography}{37}
\providecommand{\natexlab}[1]{#1}
\providecommand{\url}[1]{\texttt{#1}}
\expandafter\ifx\csname urlstyle\endcsname\relax
  \providecommand{\doi}[1]{doi: #1}\else
  \providecommand{\doi}{doi: \begingroup \urlstyle{rm}\Url}\fi

\bibitem[Albini et~al.(2020)Albini, Lertvittayakumjorn, Rago, and Toni]{DAX}
Emanuele Albini, Piyawat Lertvittayakumjorn, Antonio Rago, and Francesca Toni.
\newblock {DAX:} deep argumentative explanation for neural networks.
\newblock \emph{CoRR}, abs/2012.05766, 2020.

\bibitem[Atkinson et~al.(2017)Atkinson, Baroni, Giacomin, Hunter, Prakken, Reed, Simari, Thimm, and Villata]{AImagazine17}
Katie Atkinson, Pietro Baroni, Massimiliano Giacomin, Anthony Hunter, Henry Prakken, Chris Reed, Guillermo~Ricardo Simari, Matthias Thimm, and Serena Villata.
\newblock Towards artificial argumentation.
\newblock \emph{{AI} Mag.}, 38\penalty0 (3):\penalty0 25--36, 2017.

\bibitem[Ayoobi et~al.(2021)Ayoobi, Cao, Verbrugge, and Verheij]{Ay:AABL}
H. Ayoobi, M. Cao, R. Verbrugge, and B. Verheij.
\newblock Argue to learn: Accelerated argumentation-based learning.
\newblock In \emph{20th IEEE International Conference on Machine Learning and Applications (ICMLA)}, 2021.

\bibitem[Ayoobi et~al.(2023)Ayoobi, Potyka, and Toni]{AyoobiPT23}
Hamed Ayoobi, Nico Potyka, and Francesca Toni.
\newblock {SpArX}: Sparse argumentative explanations for neural networks.
\newblock In \emph{European Conference on Artificial Intelligence ({ECAI})}, pages 149--156. {IOS} Press, 2023.

\bibitem[Bridle(1990)]{softmax}
John~S. Bridle.
\newblock Probabilistic interpretation of feedforward classification network outputs, with relationships to statistical pattern recognition.
\newblock In \emph{Neurocomputing}, pages 227--236, Berlin, Heidelberg, 1990. Springer Berlin Heidelberg.

\bibitem[Chen et~al.(2019)Chen, Li, Tao, Barnett, Rudin, and Su]{chen2019looks}
Chaofan Chen, Oscar Li, Daniel Tao, Alina Barnett, Cynthia Rudin, and Jonathan~K Su.
\newblock This looks like that: deep learning for interpretable image recognition.
\newblock \emph{Advances in neural information processing systems}, 32, 2019.

\bibitem[Cyras et~al.(2021)Cyras, Rago, Albini, Baroni, and Toni]{argXAIsurvey}
Kristijonas Cyras, Antonio Rago, Emanuele Albini, Pietro Baroni, and Francesca Toni.
\newblock Argumentative {XAI:} {A} survey.
\newblock In \emph{Proceedings of the Thirtieth International Joint Conference on Artificial Intelligence, IJCAI 2021}, 2021.

\bibitem[Deng et~al.(2009)Deng, Dong, Socher, Li, Li, and Fei-Fei]{imageNet}
Jia Deng, Wei Dong, Richard Socher, Li-Jia Li, Kai Li, and Li Fei-Fei.
\newblock {ImageNet}: A large-scale hierarchical image database.
\newblock In \emph{2009 IEEE Conference on Computer Vision and Pattern Recognition}, pages 248--255, 2009.

\bibitem[Deng(2012)]{mnist}
Li Deng.
\newblock The {MNIST} database of handwritten digit images for machine learning research.
\newblock \emph{IEEE Signal Processing Magazine}, 29\penalty0 (6):\penalty0 141--142, 2012.

\bibitem[Goyal et~al.(2019)Goyal, Wu, Ernst, Batra, Parikh, and Lee]{goyal2019counterfactual}
Yash Goyal, Ziyan Wu, Jan Ernst, Dhruv Batra, Devi Parikh, and Stefan Lee.
\newblock Counterfactual visual explanations.
\newblock In \emph{ICML}. PMLR, 2019.

\bibitem[He et~al.(2016)He, Zhang, Ren, and Sun]{resnet}
Kaiming He, Xiangyu Zhang, Shaoqing Ren, and Jian Sun.
\newblock Deep residual learning for image recognition.
\newblock In \emph{2016 IEEE Conference on Computer Vision and Pattern Recognition (CVPR)}, pages 770--778, 2016.

\bibitem[Hendrycks and Gimpel(2023)]{gelu}
Dan Hendrycks and Kevin Gimpel.
\newblock Gaussian error linear units (gelus), 2023.

\bibitem[Hoffmann et~al.(2021)Hoffmann, Fanconi, Rade, and Kohler]{hoffmann2021looks}
Adrian Hoffmann, Claudio Fanconi, Rahul Rade, and Jonas Kohler.
\newblock This looks like that... does it? shortcomings of latent space prototype interpretability in deep networks.
\newblock \emph{arXiv preprint arXiv:2105.02968}, 2021.

\bibitem[Hu et~al.(2017)Hu, Andreas, Rohrbach, Darrell, and Saenko]{SHAPES}
Ronghang Hu, Jacob Andreas, Marcus Rohrbach, Trevor Darrell, and Kate Saenko.
\newblock Learning to reason: End-to-end module networks for visual question answering.
\newblock In \emph{2017 IEEE International Conference on Computer Vision (ICCV)}, pages 804--813, 2017.

\bibitem[Jang et~al.(2017)Jang, Gu, and Poole]{gumbelSoftmax}
Eric Jang, Shixiang Gu, and Ben Poole.
\newblock Categorical reparameterization with gumbel-softmax.
\newblock In \emph{5th International Conference on Learning Representations, {ICLR} 2017, Toulon, France, April 24-26, 2017, Conference Track Proceedings}. OpenReview.net, 2017.

\bibitem[Jo and Bengio(2017)]{jo2017measuring}
Jason Jo and Yoshua Bengio.
\newblock Measuring the tendency of cnns to learn surface statistical regularities.
\newblock \emph{arXiv preprint arXiv:1711.11561}, 2017.

\bibitem[Kim et~al.(2021)Kim, Kim, Seo, and Yoon]{xprotonet}
Eunji Kim, Siwon Kim, Minji Seo, and Sungroh Yoon.
\newblock Xprotonet: Diagnosis in chest radiography with global and local explanations.
\newblock In \emph{2021 IEEE/CVF Conference on Computer Vision and Pattern Recognition (CVPR)}, pages 15714--15723, 2021.

\bibitem[Krause et~al.(2013)Krause, Stark, Deng, and Fei-Fei]{CARS}
Jonathan Krause, Michael Stark, Jia Deng, and Li Fei-Fei.
\newblock {3D} object representations for fine-grained categorization.
\newblock In \emph{2013 IEEE International Conference on Computer Vision Workshops}, pages 554--561, 2013.

\bibitem[Krizhevsky(2009)]{cifar10}
Alex Krizhevsky.
\newblock Learning multiple layers of features from tiny images.
\newblock 2009.

\bibitem[LeCun et~al.(2015)LeCun, Bengio, and Hinton]{lecun2015deep}
Yann LeCun, Yoshua Bengio, and Geoffrey Hinton.
\newblock Deep learning.
\newblock \emph{nature}, 521\penalty0 (7553):\penalty0 436--444, 2015.

\bibitem[Loshchilov and Hutter(2019)]{adamw}
Ilya Loshchilov and Frank Hutter.
\newblock Decoupled weight decay regularization, 2019.

\bibitem[Lundberg and Lee(2017)]{SHAP}
Scott~M Lundberg and Su-In Lee.
\newblock A unified approach to interpreting model predictions.
\newblock In \emph{NeurIPS}, 2017.

\bibitem[Nauta et~al.(2021)Nauta, van Bree, and Seifert]{NautaBS21}
Meike Nauta, Ron van Bree, and Christin Seifert.
\newblock Neural prototype trees for interpretable fine-grained image recognition.
\newblock In \emph{{IEEE} Conference on Computer Vision and Pattern Recognition ({CVPR} 2021)}, pages 14933--14943. Computer Vision Foundation / {IEEE}, 2021.

\bibitem[Nauta et~al.(2023)Nauta, Schlötterer, van Keulen, and Seifert]{Nauta23}
Meike Nauta, Jörg Schlötterer, Maurice van Keulen, and Christin Seifert.
\newblock {PIP-Net}: Patch-based intuitive prototypes for interpretable image classification.
\newblock In \emph{Proceedings of the IEEE/CVF Conference on Computer Vision and Pattern Recognition (CVPR 2023)}. Computer Vision Foundation / {IEEE}, 2023.

\bibitem[Nickparvar(2021)]{brainMRIdataset}
Msoud Nickparvar.
\newblock Brain tumor mri dataset, 2021.

\bibitem[Potyka(2021)]{Potyka_21}
Nico Potyka.
\newblock Interpreting neural networks as quantitative argumentation frameworks.
\newblock In \emph{Proceedings of the Thirty-Third {AAAI} Conference on Artificial Intelligence, (AAAI-21)}, 2021.

\bibitem[Ribeiro et~al.(2016)Ribeiro, Singh, and Guestrin]{LIME_2016should}
Marco~Tulio Ribeiro, Sameer Singh, and Carlos Guestrin.
\newblock "why should i trust you?" explaining the predictions of any classifier.
\newblock In \emph{Proceedings of the 22nd ACM SIGKDD international conference on knowledge discovery and data mining}, pages 1135--1144, 2016.

\bibitem[Rudin(2019)]{Rudin19}
Cynthia Rudin.
\newblock Stop explaining black box machine learning models for high stakes decisions and use interpretable models instead.
\newblock \emph{Nat. Mach. Intell.}, 1\penalty0 (5):\penalty0 206--215, 2019.

\bibitem[Rymarczyk et~al.(2021)Rymarczyk, Struski, Tabor, and Zielinski]{RymarczykST021}
Dawid Rymarczyk, Lukasz Struski, Jacek Tabor, and Bartosz Zielinski.
\newblock {ProtoPShare}: Prototypical parts sharing for similarity discovery in interpretable image classification.
\newblock In \emph{{SIGKDD} Conference on Knowledge Discovery and Data Mining ({KDD})}, pages 1420--1430. {ACM}, 2021.

\bibitem[Rymarczyk et~al.(2022)Rymarczyk, Struski, G\'{o}rszczak, Lewandowska, Tabor, and Zieli\'{n}ski]{protoPool}
Dawid Rymarczyk, \L{}ukasz Struski, Micha\l{} G\'{o}rszczak, Koryna Lewandowska, Jacek Tabor, and Bartosz Zieli\'{n}ski.
\newblock Interpretable image classification with differentiable prototypes assignment.
\newblock In \emph{Computer Vision – ECCV 2022: 17th European Conference, Tel Aviv, Israel, October 23–27, 2022, Proceedings, Part XII}, page 351–368, Berlin, Heidelberg, 2022. Springer-Verlag.

\bibitem[Sattarzadeh et~al.(2021)Sattarzadeh, Sudhakar, Plataniotis, et~al.]{integratedcam}
Sam Sattarzadeh, Mahesh Sudhakar, Konstantinos~N Plataniotis, et~al.
\newblock Integrated grad-cam: Sensitivity-aware visual explanation of deep convolutional networks via integrated gradient-based scoring.
\newblock In \emph{IEEE Int. Conf. on Acoustics, Speech and Signal Processing (ICASSP)}, 2021.

\bibitem[Shrikumar et~al.(2017)Shrikumar, Greenside, and Kundaje]{deeplift}
Avanti Shrikumar, Peyton Greenside, and Anshul Kundaje.
\newblock Learning important features through propagating activation differences.
\newblock In \emph{ICML}, 2017.

\bibitem[Snell et~al.(2017)Snell, Swersky, and Zemel]{snell2017prototypical}
Jake Snell, Kevin Swersky, and Richard Zemel.
\newblock Prototypical networks for few-shot learning.
\newblock \emph{Advances in neural information processing systems}, 30, 2017.

\bibitem[Stallkamp et~al.(2012)Stallkamp, Schlipsing, Salmen, and Igel]{gtsrb}
J. Stallkamp, M. Schlipsing, J. Salmen, and C. Igel.
\newblock Man vs. computer: Benchmarking machine learning algorithms for traffic sign recognition.
\newblock \emph{Neural Networks}, 32:\penalty0 323--332, 2012.
\newblock Selected Papers from IJCNN 2011.

\bibitem[Sukpanichnant et~al.(2021)Sukpanichnant, Rago, Lertvittayakumjorn, and Toni]{LRPArgExpl21}
Purin Sukpanichnant, Antonio Rago, Piyawat Lertvittayakumjorn, and Francesca Toni.
\newblock Neural {QBAFs}: Explaining neural networks under lrp-based argumentation frameworks.
\newblock In \emph{AIxIA 2021 - Advances in Artificial Intelligence - 20th International Conference of the Italian Association for Artificial Intelligence, Virtual Event, December 1-3, 2021, Revised Selected Papers}, pages 429--444. Springer, 2021.

\bibitem[Wah et~al.(2011)Wah, Branson, Welinder, Perona, and Belongie]{CUB_200_2011}
C. Wah, S. Branson, P. Welinder, P. Perona, and S. Belongie.
\newblock {Caltech-UCSD Birds-200-2011 (CUB-200-2011)}.
\newblock Technical Report CNS-TR-2011-001, California Institute of Technology, 2011.

\bibitem[Xiao et~al.(2017)Xiao, Rasul, and Vollgraf]{fashionMNIST}
Han Xiao, Kashif Rasul, and Roland Vollgraf.
\newblock Fashion-{MNIST}: a novel image dataset for benchmarking machine learning algorithms.
\newblock \emph{ArXiv}, abs/1708.07747, 2017.

\end{thebibliography}
}


\end{document}